\documentclass[graybox]{svmult}

\usepackage{mathptmx}       
\usepackage{helvet}         
\usepackage{courier}        
\usepackage{type1cm}        
\usepackage{makeidx}         
\usepackage{graphicx}        
\usepackage{multicol}        
\usepackage[bottom]{footmisc}

\makeindex             

\begin{document}

\title*{Convergence Analysis of Backpropagation Algorithm for Designing an Intelligent System for Sensing Manhole Gases}
\titlerunning{Convergence Analysis of Backpropagation Algorithm}    
\author{Varun Kumar Ojha and Paramartha Dutta and Atal Chaudhuri and Hiranmay Saha}
\institute{Varun Kumar Ojha \at
    Dept. of Computer Science \& Eng., Jadavpur University, India and IT4Innovations, VSB Technical University of Ostrava, Czech Republic, \email{varun.kumar.ojha@vsb.cz}
\and  Paramartha Dutta \at Dept. of Computer \& System Sciences, Visva-Bharati University, India, \email{paramartha.dutta@gmail.com} \and Atal Chaudhuri \at Dept. of Computer Science \& Eng., Jadavapur University, India, \email{atalc23@gmail.com} \and Hiranmay Saha \at Centre of Excellence for Green Energy \& Sensors System, Indian Institute of Engineering Science and Technology, India, s\email{sahahiran@gmail.com}}
%
\maketitle

\abstract*{Human fatalities are reported due to the excessive proportional presence of hazardous gas components in manhole, such as Hydrogen Sulfide, Ammonia, Methane, Carbon Dioxide, Nitrogen Oxide, Carbon Monoxide, etc. Hence, predetermination of these gases is imperative. A neural network (NN) based intelligent sensory system is proposed for the avoidance of such fatalities. Backpropagation (BP) was applied for the supervised training of the neural network. A Gas sensor array consists of many sensor elements was employed for the sensing manhole gases. Sensors in the sensor array are responsible for sensing their target gas components only. Therefore, the presence of multiple gases results in cross sensitivity. The cross sensitivity is a crucial issue to this problem and it is viewed as pattern recognition and noise reduction problem. Various performance parameters and complexity of the problem influences NN training. In present chapter the performance of BP algorithm on such a real life application problem was comprehensively studied, compared and contrasted with the several other hybrid intelligent approaches both, in theoretical and in statistical sense.}

\abstract{Human fatalities are reported due to the excessive proportional presence of hazardous gas components in manhole, such as Hydrogen Sulfide, Ammonia, Methane, Carbon Dioxide, Nitrogen Oxide, Carbon Monoxide, etc. Hence, predetermination of these gases is imperative. A neural network (NN) based intelligent sensory system is proposed for the avoidance of such fatalities. Backpropagation (BP) was applied for the supervised training of the neural network. A Gas sensor array consists of many sensor elements was employed for the sensing manhole gases. Sensors in the sensor array are responsible for sensing their target gas components only. Therefore, the presence of multiple gases results in cross sensitivity. The cross sensitivity is a crucial issue to this problem and it is viewed as pattern recognition and noise reduction problem. Various performance parameters and complexity of the problem influences NN training. In present chapter the performance of BP algorithm on such a real life application problem was comprehensively studied, compared and contrasted with the several other hybrid intelligent approaches both, in theoretical and in statistical sense. ~}

\section{Introduction}
\label{sec:intro}
Computational Intelligence (CI) offers solution to almost every real life problem it encounters. In present chapter, we resort to using CI approach to offer a design of an intelligent sensory system (ISS) for the detection of manhole gas mixture. The manhole gas mixture detection problem is treated as a pattern recognition/noise reduction problem. In the past few years, neural network (NN) has been proved as powerful tool for machine learning application in various fields. In present chapter, we use backpropagation (BP) NN technique for the design of the said ISS. The central theme of the chapter is to present a comprehensive performance study on BP algorithm used for designing such ISS.

Decomposition of wastage and sewage into sewer pipeline leads to formation of toxic gaseous mixture often known as manhole gas mixture that usually contains toxic gases such as Hydrogen Sulphide ($H_2S$), Ammonia ($NH_3$), Methane ($CH_4$), Carbon Dioxide ($CO_2$), Nitrogen Oxide ($NO_x$), etc., \cite{SewerGas,Lewis,SewerGas1}. Often human fatalities occurs due to the presence of excessive proportion of the mentioned toxic gases in manholes. Persons, who have the responsibilities for the maintenance and cleaning of sewer pipeline are in need of a compact instrument that may predetermine the safeness of the manhole. In the recent past several instances of deaths, including municipality labourers, are reported due to toxic gas exposures \cite{NIOSH,theHinduMarch2014,toiMarch2014,theHinduApril2014,theHinduApril52014}. We have investigated the commercially available gas sensor tools. We found that the commercially available gas detectors are insufficient in sensing all the aforementioned gases as a single compact unit and the cross sensitive in the response is the basic problem associated with these sensor units. 

A brief literature survey is provided in section \ref{subsec:litrature} followed by a concise report on the contribution of the present research work in section \ref{subsec:recentWork}. Readers may find discussion on design issues of the proposed intelligent sensory system (ISS) in section \ref{sub:iss}, a discussion on data samples formation processes followed by the crucial issue of the cross sensitivity in sections \ref{subsec:datacollection} and \ref{subsec:crosssensitivity} respectively. Section \ref{sec:NN} provides brief discussion on NN configuration, training pattern and supervised BP algorithm. Performance of BP algorithm on a real application is central subject of this chapter offered in section \ref{sec:PA}. Finally, the results and conclusion is offered in section \ref{sec:res} and \ref{Conclusions} respectively.

\section{Mechanisms}
\label{sec:MaM}
The present section provides detailed discussion on various materials and methods acquired in the design and development of the ISS. Section explains the design issues of an intelligent gas detection system followed by the data collection and data preparation technique.

\subsection{A Brief Literature Survey}
\label{subsec:litrature}
In the past few years several work have been reported on electronic nose (E-NOSE) and gas detection. We have done an exhaustive survey. We may appreciate the effort by Li \textit{et al.} \cite{Jun} for his contribution in developing a NN based mixed gases ($NO_x$, and $CO$) measurement system. Sirvastava \textit{et al.} \cite{Sirvastava-ga,Sirvastava-enose} have proposed a design on intelligent E-NOSE system using BP and neuro-genetic approach. A pattern recognition technique based on wallet transformation for gas mixture analysis using single tin oxide sensor was presented by Llobet \textit{et al.} \cite{Eduard}. Liu \textit{et al.} \cite{Junhua-Liu} addressed a genetic NN algorithm to recognize patterns of the mixed gases of three components using infrared gas sensor. Tsirigotis \textit{et al.} \cite{Georgios} illustrated a NN based recognition system for $CO$ and $NH_3$ gases using metallic oxide gas sensor array (GSA). Lee \textit{et al.} \cite{Dae} illustrated the uses micro GSA with NN for recognizing combustible leakage gases. Ambard \textit{et al.} \cite{Maxim} demonstrated the use of NN for the gas discrimination using a tin oxide GSA for the gases $H_2$, $CO$  and $CH_4$. Baha \textit{et al.} \cite{Hakim} illustrated a NN based technique for development of gas sensor system for sensing gases in dynamic environment. Pan \textit{et al.} \cite{Wu} have shown the application of E-NOSE in gas mixture detection. Wongchoosuka \textit{et al.} \cite{Chatchawal} have proposed a E-NOSE based on carbon nanotube-$SnO_2$ gas sensors for detection of methanol. Zhang \textit{et al.} \cite{Qian} developed a knowledge based genetic algorithms for mine mixed gas detection. Shin \textit{et al.} \cite{Won} proposed a system for estimation of hazardous gas release rate using optical sensor and NN based technique. A comprehensive studied of the above mentioned articles results the following conclusion; (i) Mostly the BP and NN based technique are used for gas detection problem for respective application areas. (ii) Mainly, two or three gas mixture detection are addressed that too the gases are those whose sensors are not cross sensitive at high extent (iii) The issue of cross sensitivity is not addressed firmly. In design of manhole gas mixture detection system, cross sensitivity due to presence of several toxic gases is vital issue. Present article firmly addressed this issue. Ojha \textit{ et al.} \cite{Ojha-bp-j1,Ojha-lr-ic,Ojha_sa_ic} presented several approaches towards solution to manhole gas detection issue. 

\subsection{Present Approach and Contribution} 
\label{subsec:recentWork}
In the present chapter, the gas detection problem was treated as a pattern recognition/noise reduction problem where a NN regressor was modelled and trained in supervised mode using the BP algorithm. A semiconductor based gas sensor array (GSA) containing distinct semiconductor type gas sensors was used to sense the presence of gases according to their concentration in manhole gas mixture. Sensed values by the sensor array were cross sensitive as multiple gases were present in the manhole gas mixture. The cross sensitivity was occurs because the gas sensors are sensitive towards non-target gases too. Our objective was to train NN regressor such that the cross sensitivity effect can be minimized. A developed ISS would help persons to be watchful against the presence of toxic gases before entering into the manholes and thereby avoiding human fatality. Various parameters of BP algorithm were tuned to extract-out best possible result. The performance of BP for its various parameters tuning was reported comprehensively. Performance of BP against various hybrid intelligent approaches such as conjugate gradient, neurogenetic (NN trained using genetic algorithm)and neuroswarm (NN trained using particle swarm optimization algorithm) is reported both in theoretical and statistical sense. 
      
\subsection{Basic Design of Intelligent Sensory System}
\label{sub:iss}
\begin{figure}[b]
\begin{center}
\includegraphics[width=0.6 \textwidth]{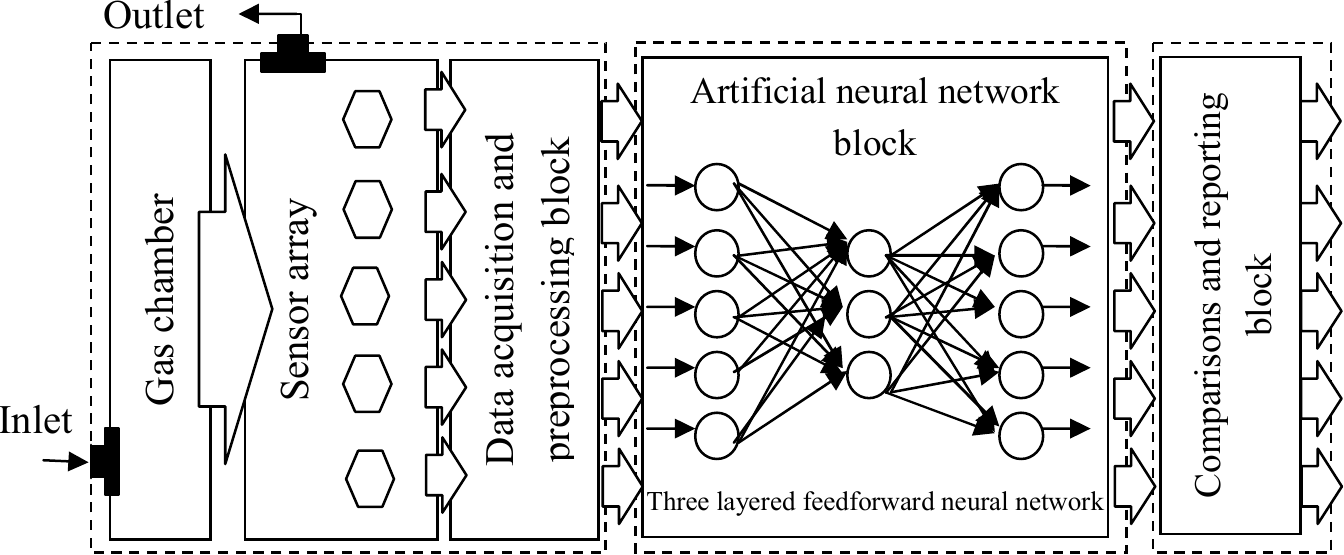}
\caption{Intelligent sensory system for manhole gas detection}
\label{fig:sysdesign}
\end{center}
\end{figure}

The design illustrated in this chapter comprised of three constituent units input unit, intelligent unit and output unit. The input unit constitutes of gas suction motor chamber, GSA and data acquisition cum data preprocessor block. The intelligent unit receives data from the input unit and after performing its computation, it sends result to output unit. Output unit presents system output in user friendly form. The gas mixture sample collected into gas mixture chamber was allowed to pass over the semiconductor based GSA. The preprocessing block receives sensed data values from the GSA and ensure that the received data values are normalized before feeding it to the NN. The output unit does the task of denormalization of the network response. It generates alarm, if any of the toxic gas components exceeds their safety limit. For the training of the network several data samples were prepared. The block diagram shown in Figure \ref{fig:sysdesign} is a lucid presentation of the above discussion.

\subsection{Semiconductor based Gas Sensor Array (GSA) and Cross Sensitivity Issue}   
\label{subsec:GSA}   
\label{subsec:crosssensitivity}  
\begin{figure}[t]
\begin{center}
\includegraphics[width=0.45 \textwidth]{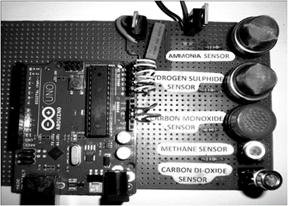}
\caption{Sensor array with data acquisition system}
\label{fig:sensorarray}
\end{center}
\end{figure}

The metal oxide semiconductor gas sensors were used to form GSA. $N$ number of distinct sensor element of $n$ gases constitutes a one dimensional GSA. The MOS sensors are basically resistance type electrical sensors. A resistance type sensor respond as change in resistance on change in the concentration of gases. The change in resistance is given as $\triangle R_s/R_0$, where the $\triangle R_s$ is the change in resistance of the MOS sensor and $R_0$ is the base resistance value \cite{Chatchawal,Dae}. A typical arrangement of a GSA is shown in Figure \ref{fig:sensorarray}. The circuitry shown in Figure \ref{fig:sensorarray} is developed in our laboratory. Ghosh \textit{et al.} \cite{GSA,PortableGSA} we have elaborately discuss the sensor array and its working principles.

Although the gas sensor elements were suppose to detect their target gases only, they showed sensitivity towards other gases too. Hence, the sensor array response was always involving cross-sensitivity effect \cite{Jun}. This indicates that the sensors responses were noisy. If we concentrate on the first and second rows in Table \ref{tab:rawdatasample}, we may appreciate the inherent cross sensitivity effect in the responses of sensors. The first and second sample in Table \ref{tab:rawdatasample}, indicates that changes in concentration of only Methane gas resulted in change of responses of all the other sensors, including the sensor earmarked for Methane. It is indicating that the prepared data sample was containing cross sensitive effect. It may also be observed that the cross-sensitivity effect was not random, rather followed some characteristics and patterns. Hence, in the operative (real world) environment the sensor responses of the GSA may not be able to use directly for the prediction of the concentration of the gases in manhole gas mixture. Therefore, to predict/forecast the level of concentration of the gases in the manhole gas mixture, we proposed to use ISS equipped with pattern recognition/noise reduction techniques that will help to filler-out noise induced on the sensors due to the cross sensitivity. 

\subsection{Data Collection Method}   
\label{subsec:datacollection}   
Data sample for experiment and NN training was prepared in several steps. In first step, information about the safety limits of the component gases found in manhole gas mixture was collected. Then distinct concentration values (level) around the safety limits of each  manhole gas was recognize. Several gas mixture samples were prepared by mixing gas components in different combination of their concentration. As an example, if we have five gases and we have recognized three concentration level of each gases, then we may mix them in 243 different combination. Hence, we may obtain 243 samples of gas mixture. When these mixture samples were allowed to pass over the semiconductor based GSA one by one in order to produced a data sample table for our experiment. A typical example of such data sample is shown in Table \ref{tab:rawdatasample}. 
\begin{table}
\begin{center}
\caption{Data sample for ISS}
\label{tab:rawdatasample}
\begin{tabular}{c | c c c c c | c c c c c}
	\hline
	& \multicolumn{5}{p{3cm}|}{\centering \textbf{Sample gas mixture (in ppm)}} & 					\multicolumn{5}{p{4cm}}{\centering \textbf{Sensor Response ($\triangle R_s/R_0$)}}\\
	\cline{2-11}
	\multicolumn{1}{c|}{\textbf{\# }}
	& \multicolumn{1}{c}{$NH_{3}$}
	& \multicolumn{1}{c}{$CO $}
	& \multicolumn{1}{c}{$H_{2}S$}
	& \multicolumn{1}{c}{$CO_{2}$}
	& \multicolumn{1}{c|}{$CH_{4}$}
	& \multicolumn{1}{c}{$NH_{3}$}
	& \multicolumn{1}{c}{$CO $}
	& \multicolumn{1}{c}{$H_{2}S$}
	& \multicolumn{1}{c}{$CO_{2}$}
	& \multicolumn{1}{c}{$CH_{4}$}\\
	\hline
	 1 & 50 & 100 & 100 & 100 & \textbf{2000} & 0.053 & 0.096 & 0.065 & 0.037 & 			0.121\\
	 2 & 50 & 100 & 100 & 100 & \textbf{5000} & \textbf{0.081} & \textbf{0.108} & \textbf{0.074} & \textbf{0.044} & \textbf{0.263}\\
	 3 & 50 & 100 & 100 & 200 & 2000 & 0.096 & 0.119 & 0.092 & 0.067 & 						0.125\\
	 4 & 50 & 100 & 200 & 200 & 5000 & 0.121 & 0.130 & 0.129 & 0.079 & 						0.274\\
	 5 & 50 & 100 & 200 & 400 & 2000 & 0.145 & 0.153 & 0.139 & 0.086 & 						0.123\\
	\hline
\end{tabular}
\end{center}
\end{table}

\subsection{Neural Network Approach}
\label{sec:NN}
As it has been already mentioned in the section \ref{sec:MaM} that the raw sensor response may not represent real world scenario accurately. Therefore, we were inclined to use NN technique to reduce noise in order to predict/represent the real world scenario with lowest possible error.
      
\subsubsection{Multi layer perceptron}
\label{subsec:mlp}
\begin{figure}[b]
\begin{center}
\includegraphics[width=0.4 \textwidth]{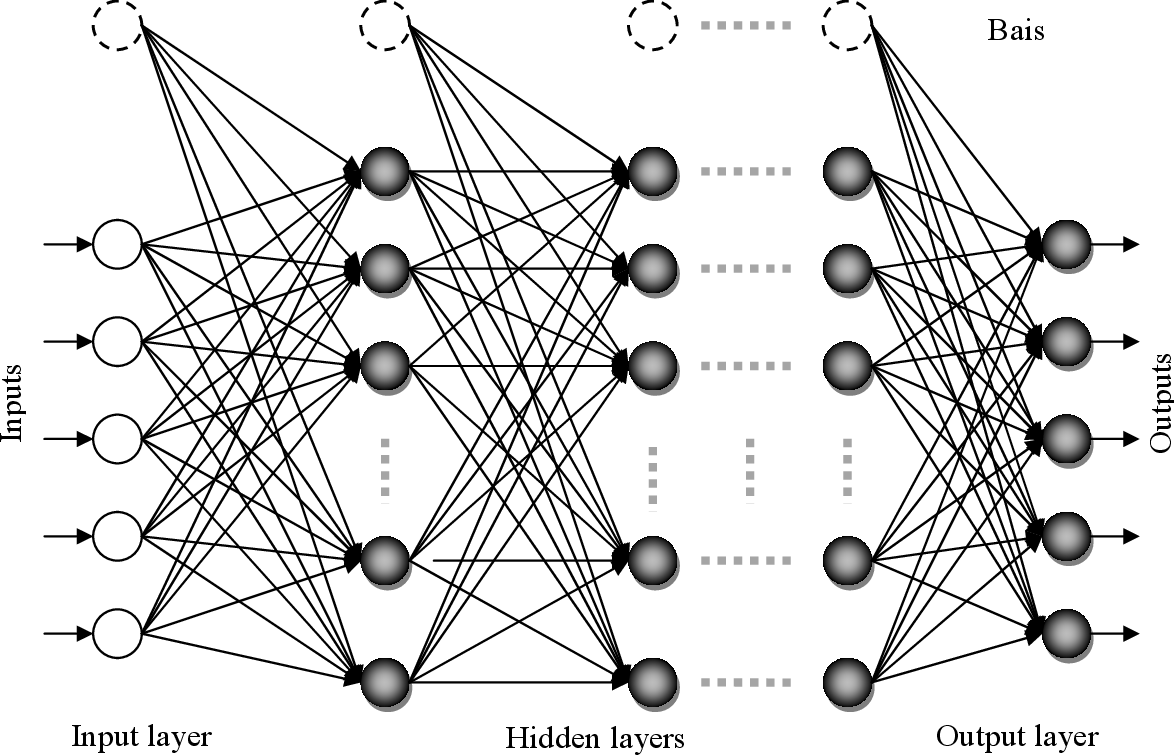}
\caption{NN architecture for five input manhole gas components}
\label{fig:nn}
\end{center}
\end{figure}

NN, \textit{``a massively parallel distributed processor that has a natural propensity for storing experiential knowledge and making it available for subsequent use"} \cite{Simon} trained using BP algorithm may offer solution to the aforementioned problem. The NN shown in Figure \ref{fig:nn} is containing $5$ input nodes, $n$ hidden nodes with $l$ layers and $5$ output nodes leading to a $5 - n  \ldots n - 5$ network configuration. The $5$ nodes in the input as well as in the output layer indicate that the system was developed for detecting $5$ gases from the gaseous mixture. A detail discussion on network configuration is provided in section \ref{subsec:nw_config}.

\subsubsection{Training Pattern} 
\label{subsec:tp}     
We acquired supervised mode of learning for the training of NN. So, training pattern constituted of input vector and target vector. It is also mentioned above that, the normalized sensor responses are given as input to the NN. So the input vector $I$ consisted of normalized values of the sensor responses. In the given data sample input vector was a five element vector, where  each element in the input vector represented a gas in the sample gas mixture. The input vector $I$ can be represented as follows:
\begin{equation}
I = [i_1, i_2, i_3, i_4, i_5]^{T}
\end{equation}

System output was presented in terms of the concentration of gases. So, the target vector $T$ was prepared using values of gas mixture sample. In the given data sample target vector was a five element vector, where each element in target vector represented a gas in the sample gas mixture. The target vector $T$ can be represented as follows:
\begin{equation}
T = [t_1, t_2, t_3, t_4, t_5]^{T}
\end{equation} 
A training sat containing input vector and target vector can be represented as per Table \ref{tab:trainingset}.
\begin{table}[b]
\begin{center}
\caption{Training set for neural network}
\label{tab:trainingset}
	\begin{tabular}{c | c c c c c | c c c c c }
	\hline
	& \multicolumn{5}{c|}{\centering \textbf{Input Vector} $I$} & \multicolumn{5}		{c}	{\centering \textbf{Target Vector} $T$ }\\
	\cline{2-11}
	\multicolumn{1}{c|}{\textbf{\# Pattern}}
	& \multicolumn{1}{c}{$i_1$}
	& \multicolumn{1}{c}{$i_2$}
	& \multicolumn{1}{c}{$i_3$}
	& \multicolumn{1}{c}{$i_4$}
	& \multicolumn{1}{c|}{$i_5$}
	& \multicolumn{1}{c}{$t_1$}
	& \multicolumn{1}{c}{$t_2$}
	& \multicolumn{1}{c}{$t_3$}
	& \multicolumn{1}{c}{$t_4$}
	& \multicolumn{1}{c}{$t_5$}\\
	\hline
	 $(I_1,T_1)$ & 0.19 & 0.35 & 0.23 &   0.13 &	0.44 & 0.01	& 0.02 & 0.02 	& 		0.02 & 0.4\\
	 $(I_2,T_2)$ & 0.29 & 0.39 & 0.27 &	0.16 &	0.96 & 0.01	& 0.02 & 0.02 & 0.02 			& 1.0  \\
     $(I_3,T_3)$ & 0.35 & 0.43 & 0.33 &	0.24 &	0.45 & 0.01 & 0.02 & 0.02 &	0.04 			& 0.4\\
	 $(I_4,T_4)$ & 0.44 & 0.47 & 0.47 &	0.28 &	1.00 & 0.01 & 0.02 & 0.04 & 0.04 			& 1.0  \\
	 $(I_5,T_5)$ & 0.52 & 0.55 & 0.51 &	0.31 &	0.45 & 0.01 & 0.02 & 0.04 & 0.08 			& 0.4\\
	\hline 
\end{tabular}
\end{center}
\end{table}

\subsubsection{The backpropagation (BP) algorithm}   
\label{subsec:bpalgo}   
\begin{figure}
\begin{center}
\includegraphics[width=0.4 \textwidth]{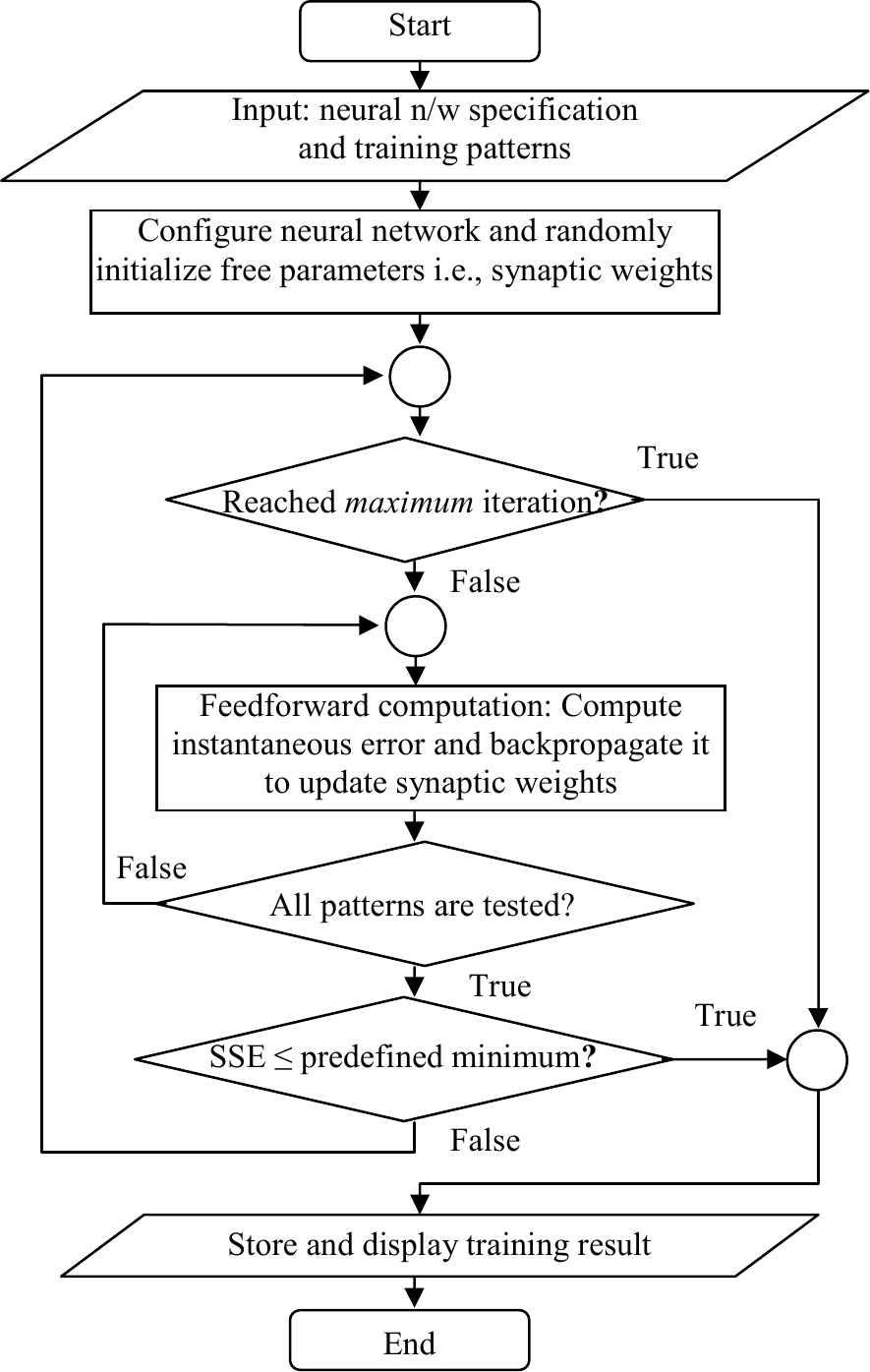}
\caption{Schematic flow chart of BP algorithm}
\label{fig:bpflowchart}
\end{center}
\end{figure}

Let us have a glimpse of BP algorithm as described by Rumelhart in \cite{Rummelhart}. BP algorithm is a form of supervised learning for multilayer NNs, also known as the generalized delta rule \cite{Rummelhart}. Error data at the output layer is back propagated to earlier ones allowing incoming weights to be updated \cite{Rummelhart,Simon}. The synaptic weight matrix $W$ can be updated as per delta rule is as: 
\begin{equation}
W(n+1) = W(n) + \triangle W(n+1),
\end{equation}
where, $n$ indicates $n^{th}$ epoch training and $\triangle W(n+1)$ is computed as
\begin{equation}
\triangle W(n+1)  =  \eta g(n+1) + m . g(n), 
\end{equation}
where, $\eta $ is learning rate, $\beta$ is the momentum factor and gradient $g(n)$ for $n^{th}$ epoch is computed as
\begin{equation}
\label{eq:grad}
g(n) = \delta_j(n)y_i(n).
\end{equation} 
The local gradient $\delta_j(n)$ is computed for both output layer and hidden layer as follows.
\begin{eqnarray}
\delta_j(n) & = & - e_j(n) \varphi(v_j(n) \hspace{0.87in} \mbox{ for \hspace{0.04in}output \hspace{0.04in}layer} \nonumber \\ 
            & = & \varphi(v_j(n) \sum\limits_k \delta_k(n) w_k{j}(n) \hspace{0.2in} \mbox{for \hspace{0.04in}hidden \hspace{0.04in}layer}  
\end{eqnarray} 
The algorithm terminates either when the sum of squared error (SSE) reached to an acceptable minimum or when the algorithm completes its maximum allowed iterations. The SSE measures the performance of BP algorithm. The SSE \cite{Sivanadam} may be computed as 
\begin{equation}
SSE = \frac{1}{2} \sum\limits_p\sum\limits_i (O_{pi} - t_{pi})^{2} \hspace{0.5in}  \forall \hspace{0.04in}p \hspace{0.04in}\& \hspace{0.04in}i, 
\end{equation}
where $O_{pi}$ and $t{pi}$ are the actual and desired outputs respectively realized at the output layer, $p$ is the input pattern vector and $i$ is the number of nodes in the output layer
A flow-diagram shown in Figure \ref{fig:bpflowchart} clearly illustrates the aforementioned BP algorithm.
 
\section{Performance Study based on Various Parameters}
\label{sec:PA}
The BP algorithm was implemented using JAVA programming language. The training of NN modeled for the said problem was provided using data sample prepared as per method indicted in Table \ref{tab:trainingset}. Thereafter, performance of the BP algorithm was observed. An algorithm used for training of NN for any particular application is said to be efficient if and only if, the SSE or mean square error (MSE) induced on the NN for given training set can be reduced to an acceptable minimum. BP algorithm is robust and popular algorithm used for the training of multilayer perceptrons (MLPs). The performance analysis presented in this section, aims to provide an insight on the strengths and weaknesses of BP algorithm used for the application problem mentioned. The performance of the BP algorithm depends on adequate choice of various parameters used in the algorithm and the complexity of the problem, the algorithm is applied on. We may not control the complexity of the problem, but we may regulate various parameters to enhance the performance of the BP algorithm. Even though the BP is widely used NN training algorithm, the several controlling parameters is one of the reasons that motivated research community to think of the alternatives of the BP algorithm. Our study, illustrates the influence of various parameters on the performance of BP algorithm.

\subsection{Parameter Tuning of BP Algorithm}
\label{subsec:pa}
The performance of BP algorithm is highly influenced by various heuristics and parameters used \cite{Rummelhart,Simon}. These include training mode, network configuration (network model), learning rate ($\eta$), momentum factor ($\beta$), initialization of free parameters (synaptic weights) i.e., initial search space ($\chi$), volume of training set and terminating criteria.

\subsubsection{Mode of Training}
\label{subsec:mode_of_training}
\begin{figure}[t]
   \centering
   \includegraphics[width=2in,height=1.5in]{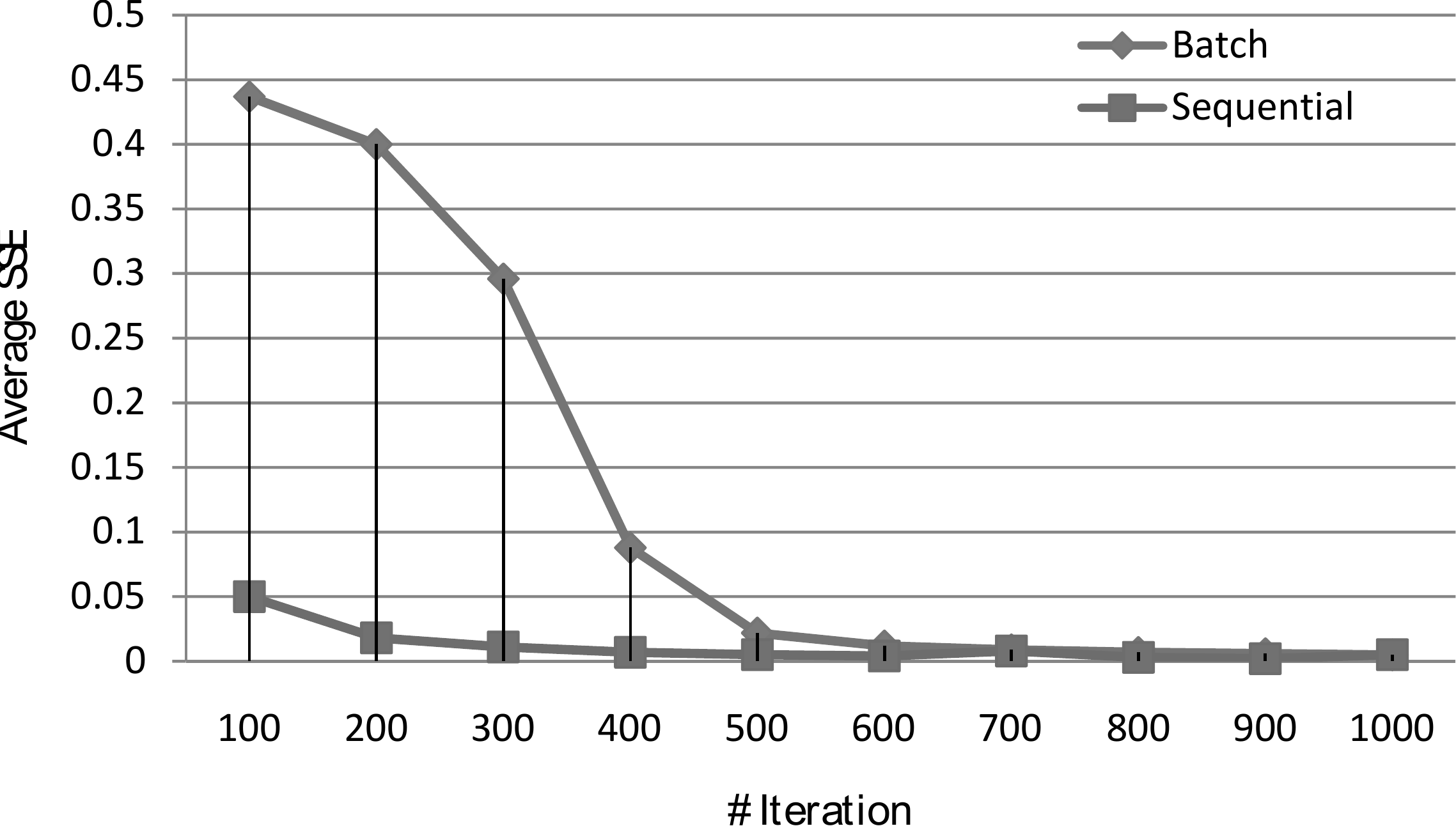} \quad \includegraphics[width=2in,height=1.5in]{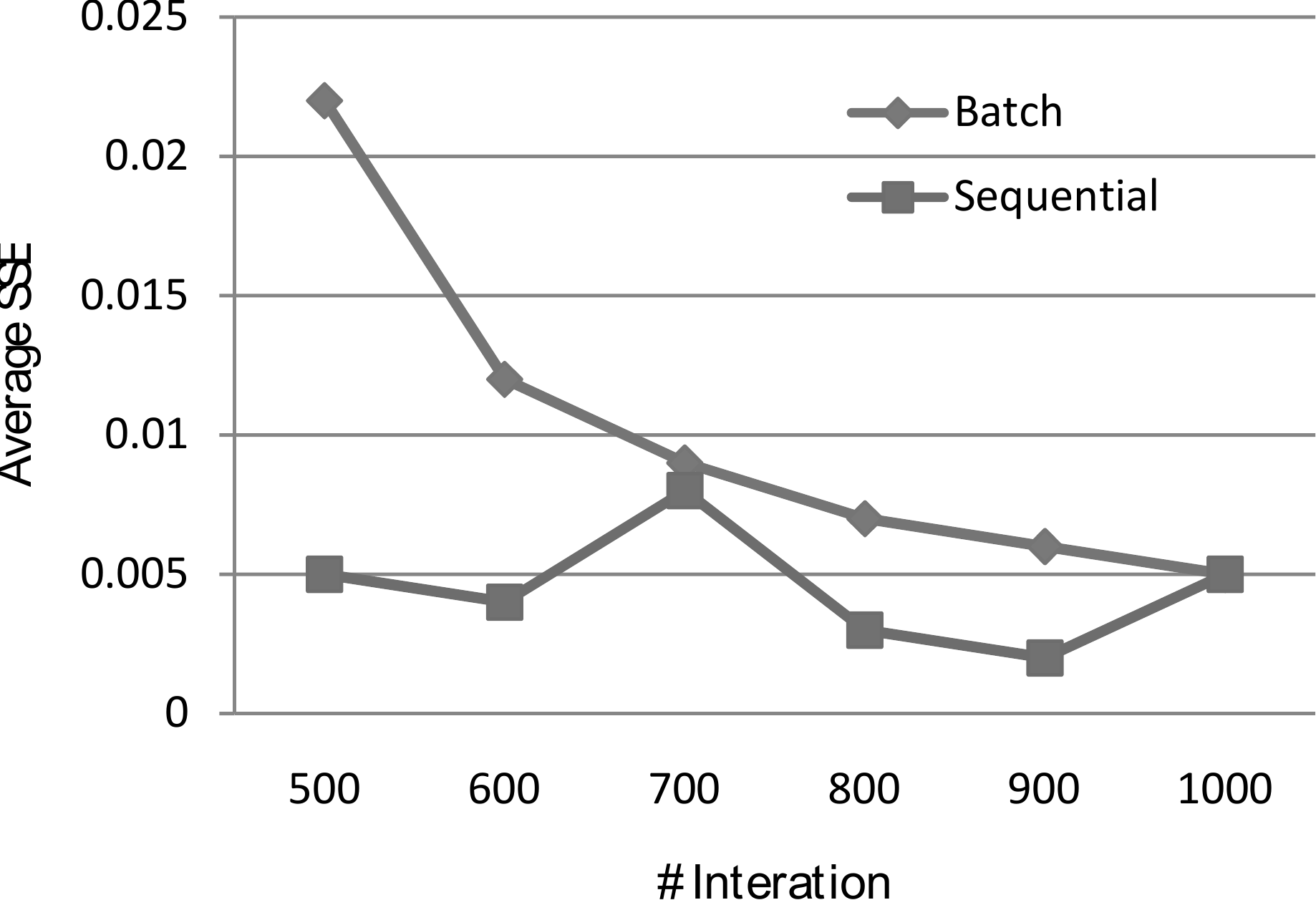}
   \caption{Convergence: Seq. vs Batch mode}
   \label{fig:mode-a}   
   \label{fig:mode-b}
   \vspace{5pt}
   \includegraphics[width=2in,height=1.5in]{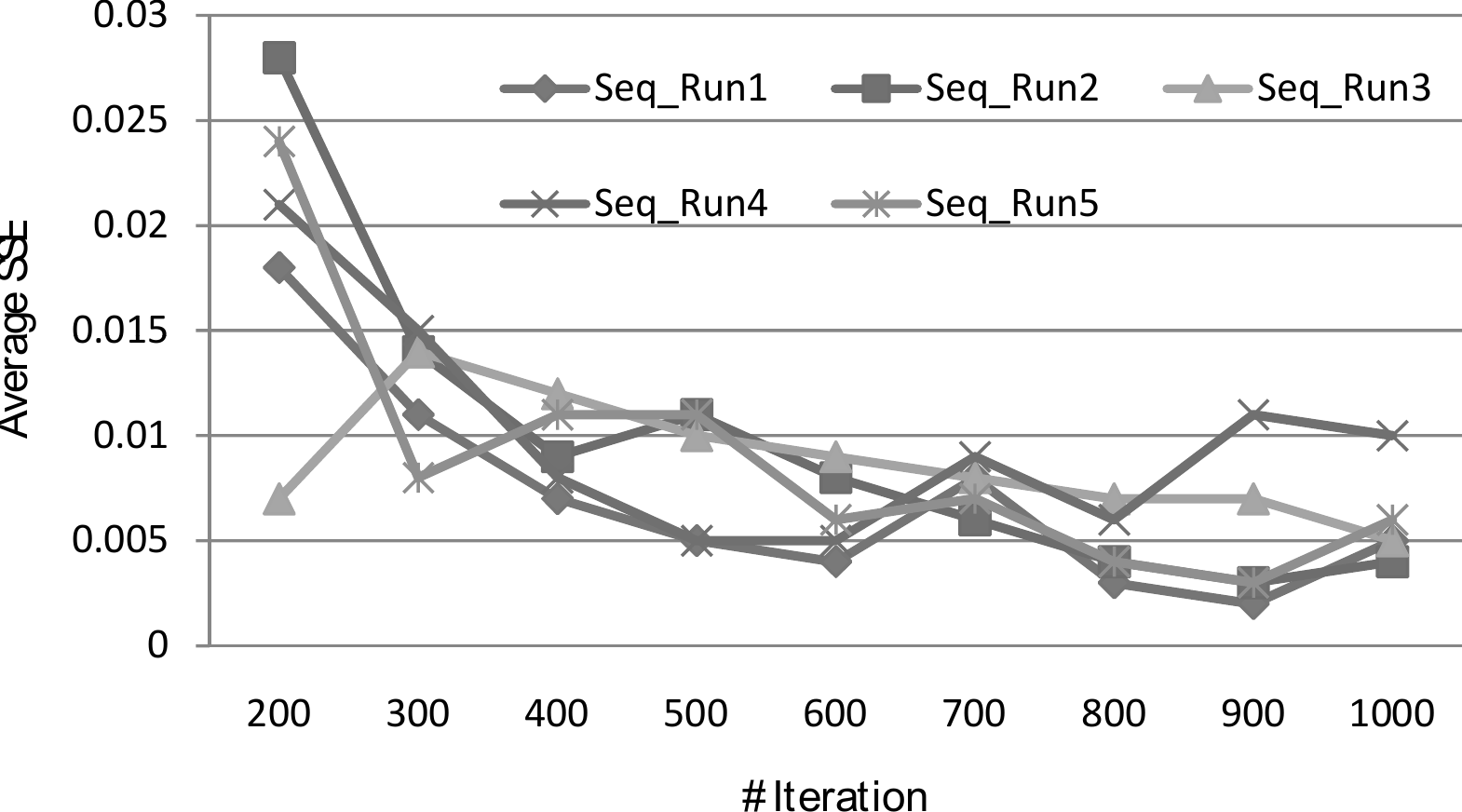} \quad  \includegraphics[width=2in,height=1.5in]{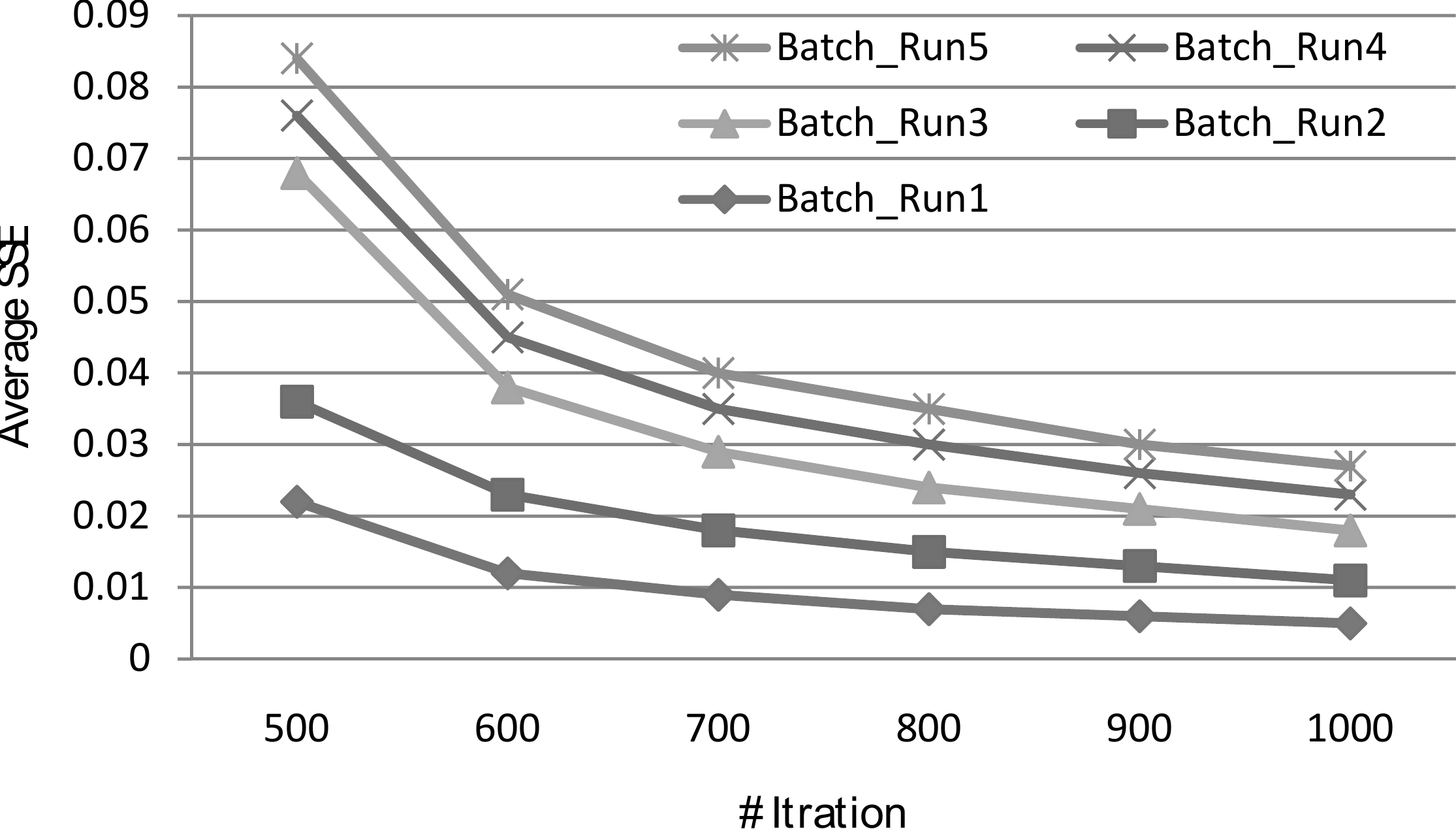}
   \caption{Convergence trajectory in (left) Seq. mode  and (right) batch mode }
   \label{fig:mode-c}
   \label{fig:mode-d}
\end{figure}

In BP algorithm, two distinguish modes of training sequential mode and batch mode can be employed to train the NN. Both of these training methods have their own advantages and disadvantages. In sequential mode, the free parameters (synaptic weights) of NN are made to adjust for each training example in the entire training set. In other words, the synaptic weight adjustment is based on instantaneous error induced on the NN for each instance of training example in the entire training set. This particular fashion of weight adjustment makes sequential mode training easy and simple to implement. Since the sequential mode training is stochastic in nature, convergence may not follow smooth trajectory, but it may avoid being stuck into local minima and may lead to global optimum if one exists. On the contrary, in batch mode, the free parameters are updated once in an epoch. An epoch training indicates that the adjustment of free parameters of NN takes place once for the entire training set. In other words, the training of NN is based on the SSE induced on the NN. Hence, gradient computation in this particular fashion of training is more accurate. Hence the convergence may be slow but may follow smooth trajectory. Figures \ref{fig:mode-a} and \ref{fig:mode-d} indicate the performance of BP algorithm based on two modes of training. Figure \ref{fig:mode-a} indicates that the convergence using batch mode training method was slower than sequential mode training. From Figures \ref{fig:mode-b} and \ref{fig:mode-c}, it may be observed that the sequential mode training method may not follow smooth trajectory, but it may tend to global optimum whereas, it is evident that in batch mode training method convergence was following smooth trajectory. 

\subsubsection{Network Configuration}
\label{subsec:nw_config}
Multi-layer perceptrons are used for solving nonlinear problems. The basic MLPs configuration consists of one input layer, one output layer with one or more hidden layer(s). Each layer may consist of one or more processing unit (neurons). The primary task in devising problems in NN is the selection of appropriate network model. Finding an optimum NN configuration is also known as structural training of NN. In network model, selection process is initiated with the selection of most basic three layer architecture. Three layer NN consists of one input layer, one hidden layer and one output layer. Number of neurons in input layer and output layer depends on the problem itself. The gas detection problem is essentially a noise reduction problem, whre the NN tries to reduce noise of signals emitting from sensors in sensor array in the presence of gas mixture. Hence, it was obvious that the number of outputs equals the number of inputs. In this application, we were designing system that may detect five gases. As there was five input signals to NN that leads to five processing units at input layer and five processing units at output layer. In three layer NN configuration, the network configuration optimization reduces to the scope of regulating number of processing units (neurons) at hidden layer. Keeping other parameters fixed to certain values, the number of nodes at hidden layer were regulated from 1 to 8 in order to observe the influence of NN configuration in the performance of BP algorithm. The parameter setup was as follows: number of iterations was set to thousand, $\eta$ to 0.5, $\beta$ to 0.1, $\chi$ $\in$ [-1.5 ,1.5], training volume to 50 training examples.
\begin{figure}[t]
  \centering
     \includegraphics[width=2in,height=1.5in]{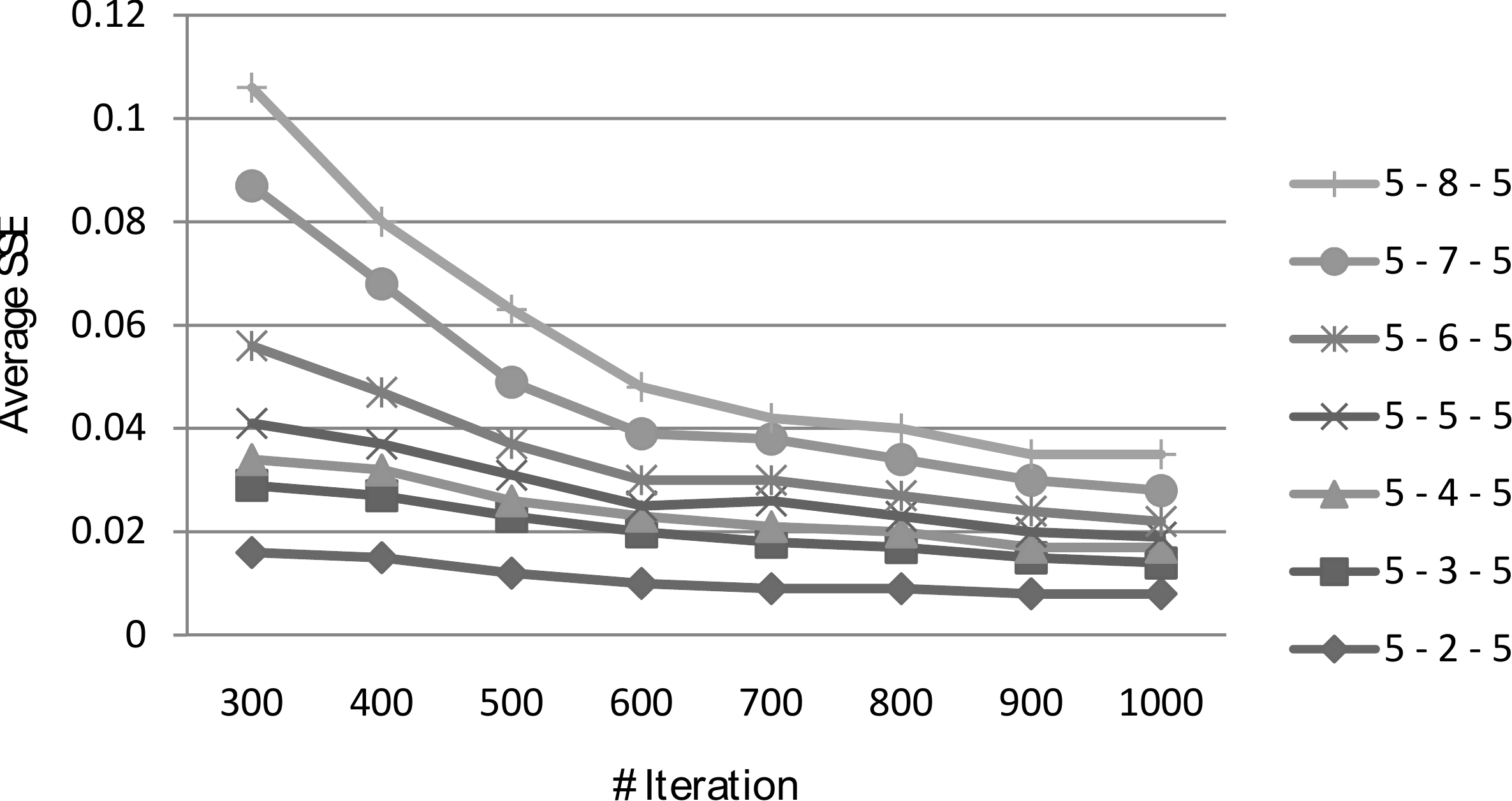} \quad \includegraphics[width=2in,height=1.5in]{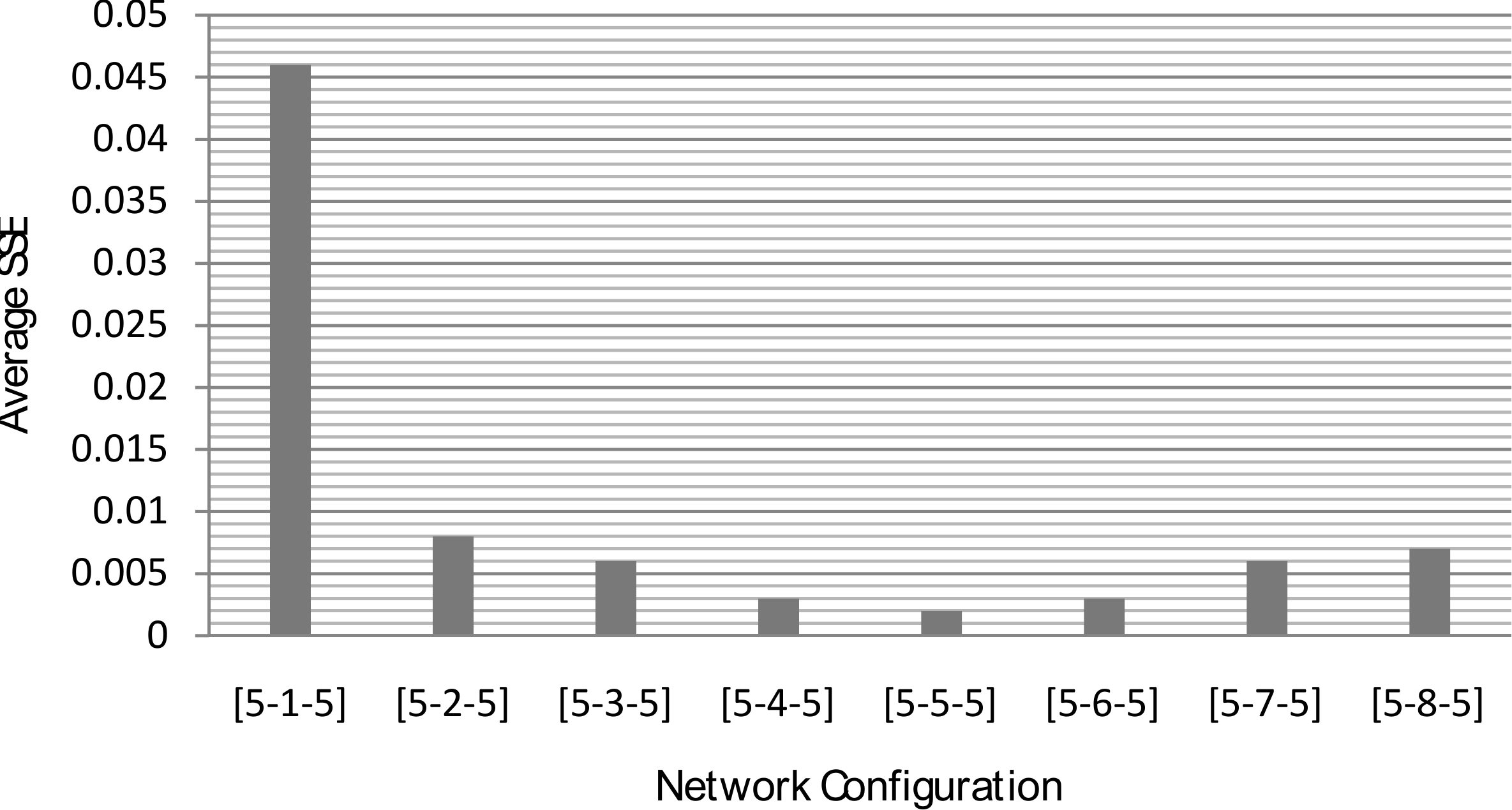}
     \caption{(left) Convergence trajectory at NN Configuration; (right) SSE at various NN Configuration.}
     \label{fig:network-a}
     \label{fig:network-b}
\end{figure}

Figure \ref{fig:network-a} demonstrates the performance of BP algorithm based on network configuration, where the horizontal axis (X-axis) indicates the change in the number of nodes at hidden layer while the vertical axis (Y-axis) indicates the average value of the SSE obtained against different network configurations. From Figure \ref{fig:network-a}, it evident that algorithm performs better with respect to the network configuration 5 - 4 - 5, where value 4 indicates the number of nodes at hidden layer. For configuration higher than 5 - 5 - 5, the performance of the algorithm was becoming poorer and poorer. Figure \ref{fig:network-b} indicates the convergence trajectory of BP algorithm for different network configuration. It may be observed that the convergence speed was slower for higher network configurations than the lower network configuration, because the higher network configuration has large number of free parameters in comparison to lower network configuration. Therefore, the higher configuration need more number of iteration to converge.

We may increase the number of hidden layers to form four or five layered NN. For the sake of simplicity, the number of nodes was kept same at each hidden layer. It has been observed that the computational complexity was directly proportional to the number of hidden layers in the network. It was obvious to bear additional computational cost if the performance of the algorithm improves for the increasing number of hidden layers. It has also been observed that the performances of algorithm and network configuration are highly sensitive to the training set volume. Performance study based on training set volume is discussed in section \ref{subsec:training_vol}.

\subsubsection{Initialization of Free Parameters (Initial Search Space - $\chi$ )}
Proper choice of $\chi$ contributes to the performance of BP algorithm. Free parameters of BP algorithm were initialized with some initial guess. The remaining parameters were kept fixed at certain values and the synaptic weights were initialized between [-0.5, 0.5] and [-2.0, 2.0] in order to monitor the influence of synaptic weights' initialization. The parameters were set as follows: the number of iteration was set to thousand, $\eta$ to 0.5, $\beta$ to 0.1, network configuration to 5 - 5 - 5, training volume to 50 training examples.
\begin{figure}[b]
  \centering
     \includegraphics[width=2in,height=1.5in]{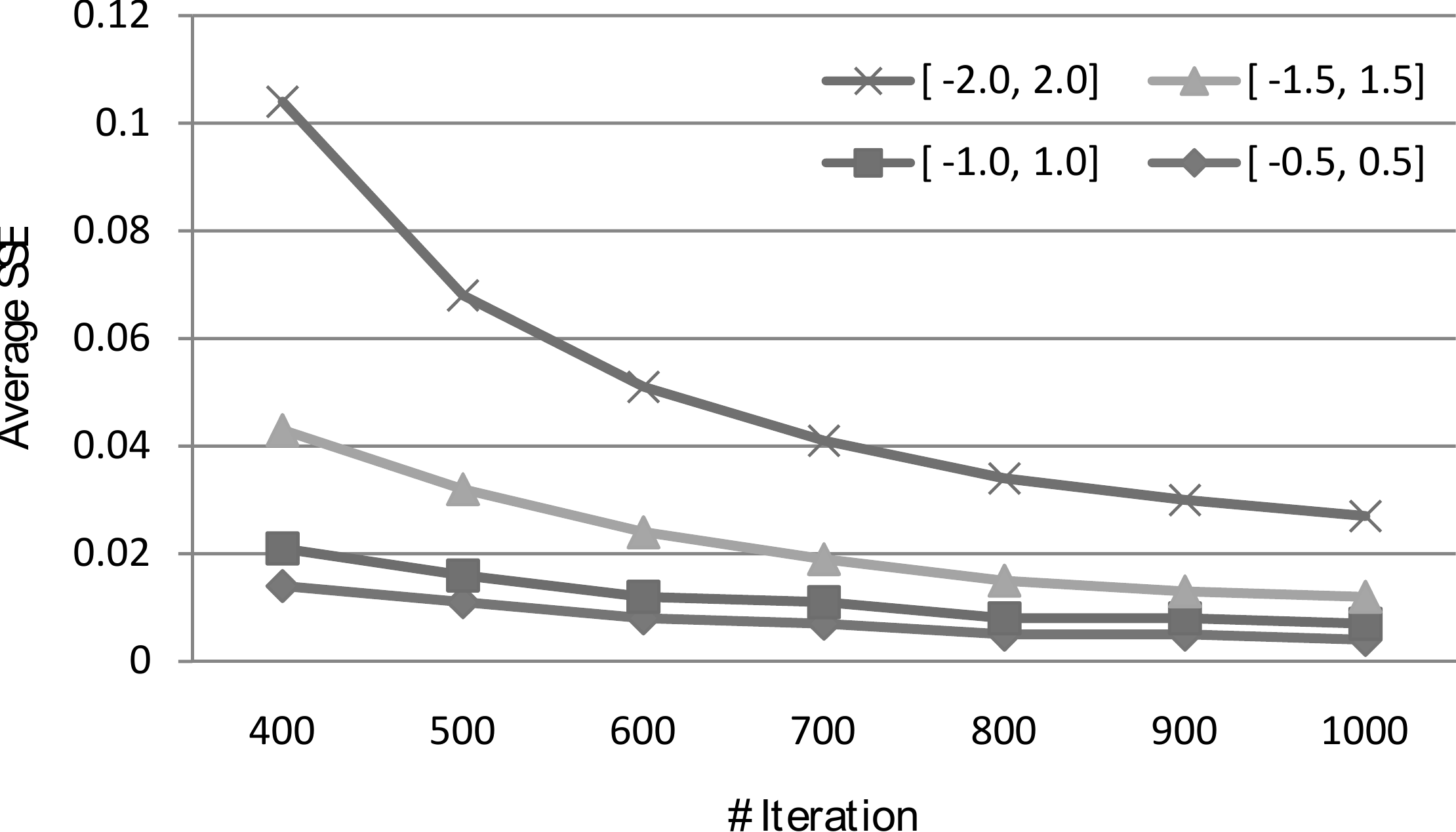} \quad \includegraphics[width=2in,height=1.5in]{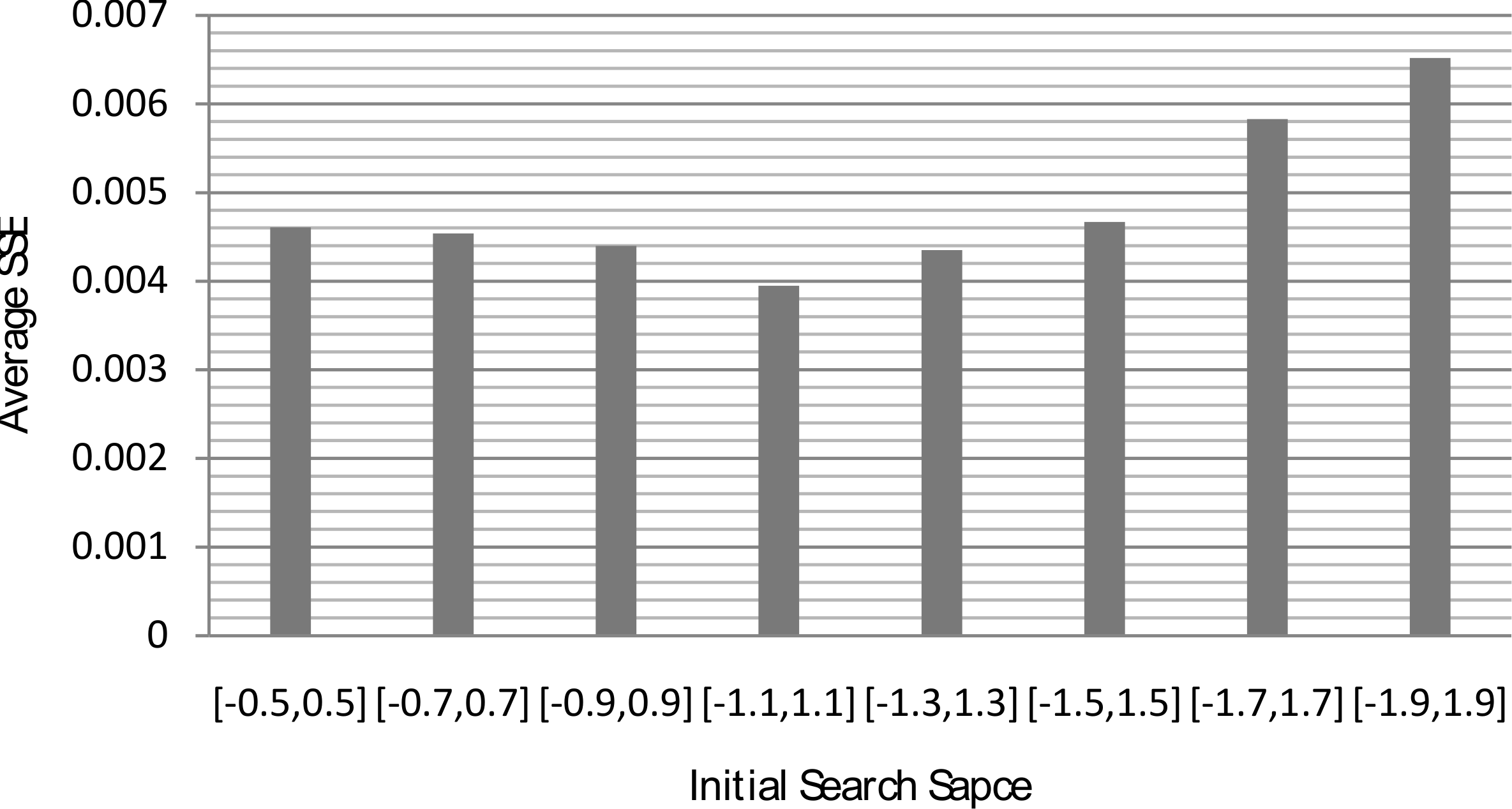}
     \caption{(left) Convergence trajectory at $\chi$; (right) SSE at various $\chi$}
     \label{fig:weights-a}
     \label{fig:weights-b}
\end{figure}

In Figure \ref{fig:weights-a}, the X-axis represents the iteration whereas the Y-axis represents average SSE. Four continuous lines in Figure \ref{fig:weights-a} indicates convergence trajectory for different initialization values. Figure \ref{fig:weights-a} demonstrate that large initial values lead to small local gradient that causes learning to be very slow. It was also observed that the learning was good somewhere between small and large initial values. In this case, $\pm 1.1$ was a good choice of $\chi$ range.

\subsubsection{Convergence Trajectory Analysis}
\begin{figure}
  \centering
    \includegraphics[width=2in,height=1.5in]{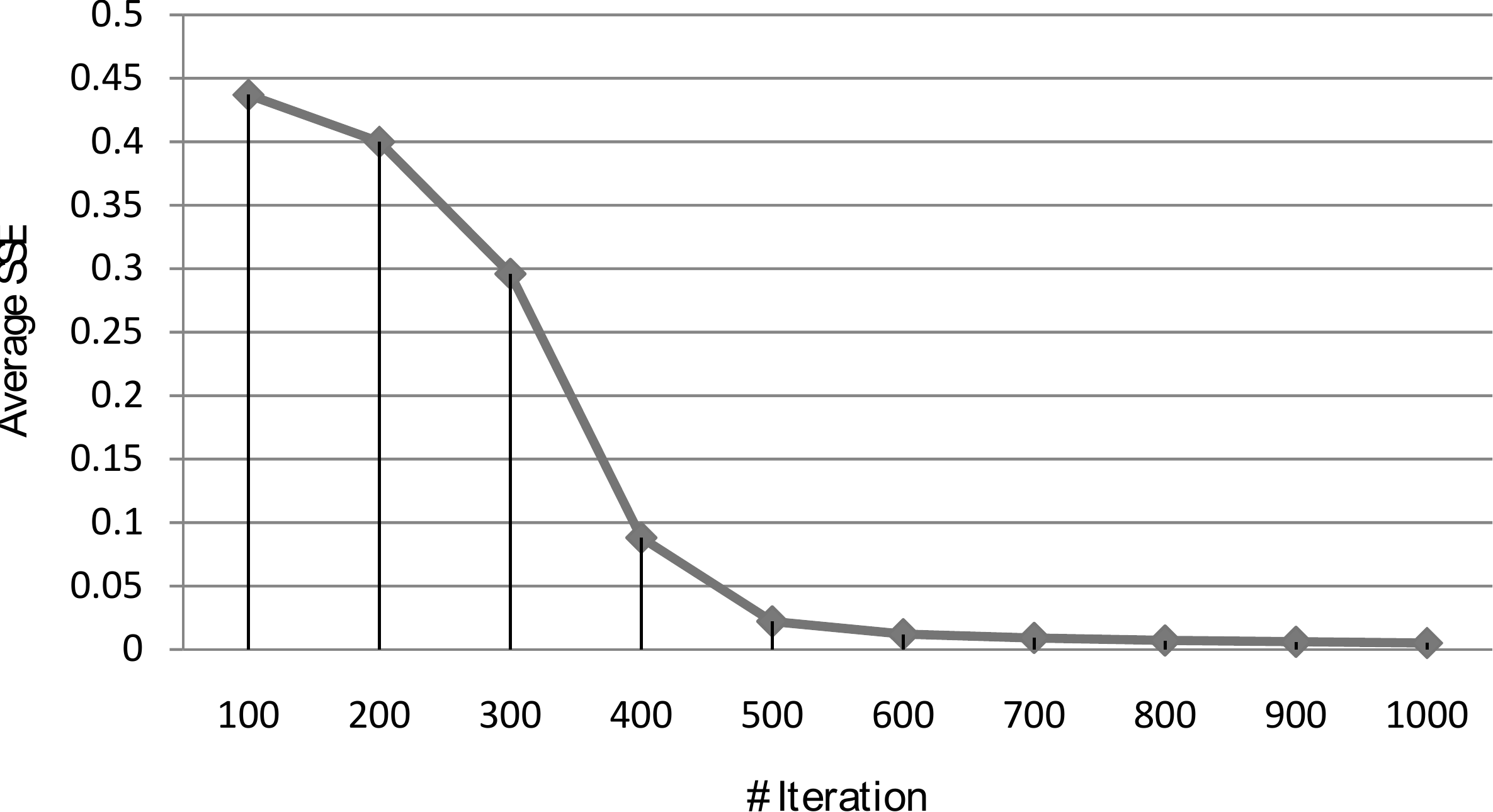} \quad   \includegraphics[width=2in,height=1.5in]{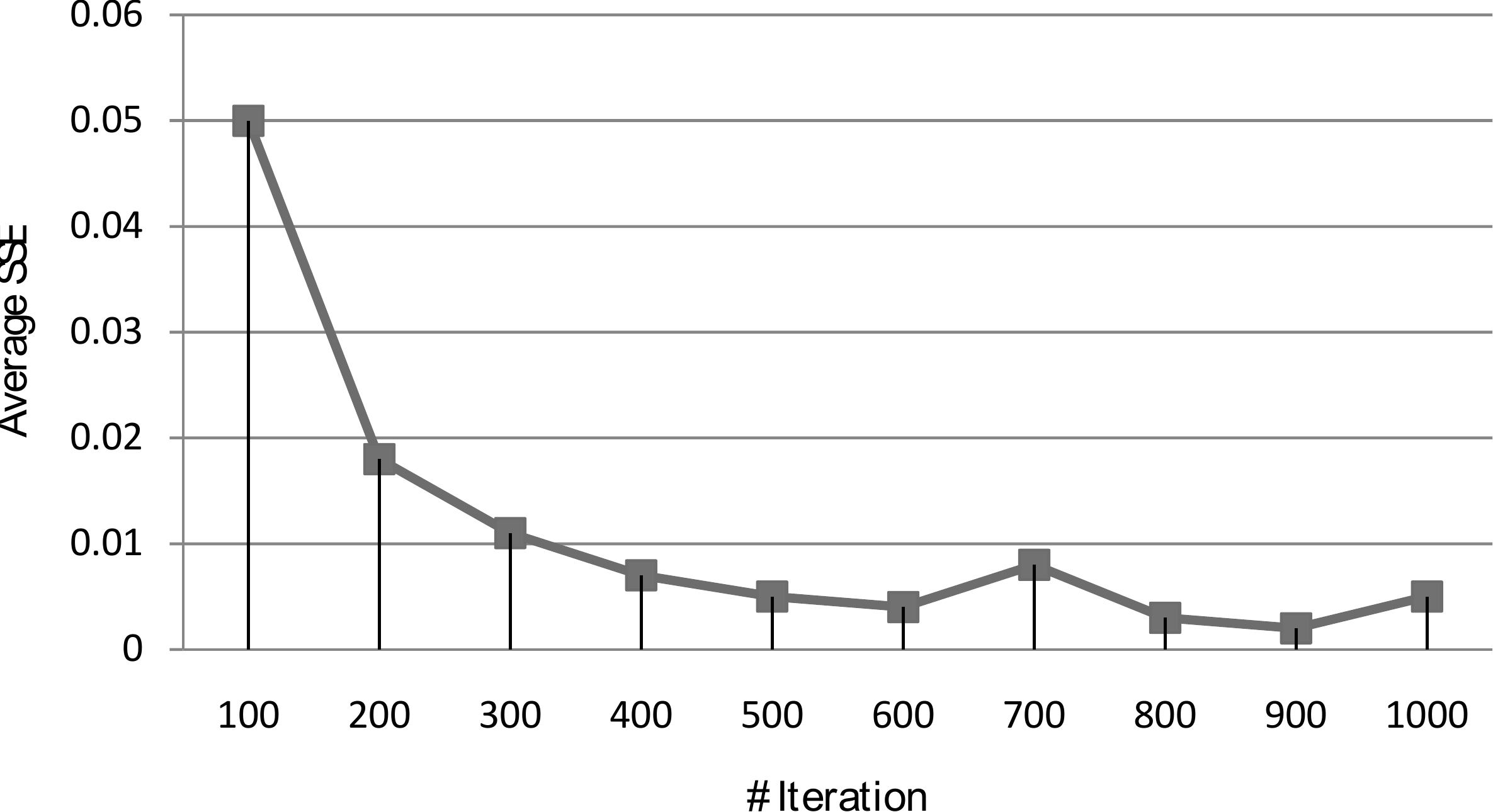}
    \caption{(left) Convergence in batch mode; (right) Convergence in sequential mode}
    \label{fig:itr-a}
    \label{fig:itr-b}
\end{figure}

In Figure \ref{fig:itr-a} the X-axis indicates the number of iteration while the Y-axis indicates the average SSE achieved. The parameters such as network configuration set at 5 - 5 - 5, number of iteration taken is 10000, $\eta$ taken is 0.5, $\beta$ taken is 0.1, training set volume is 50. Figure \ref{fig:itr-a} indicates the convergence trajectory of BP algorithm, where it may be observed that the performance of BP improves while iteration number increases. For given training set and network configuration the average SSE gets reduced to 0.005 at iteration number 1000.

\subsubsection{Learning Rate and Momentum Factor}
\begin{figure}[t]
  \centering
    \includegraphics[width=2in,height=1.5in]{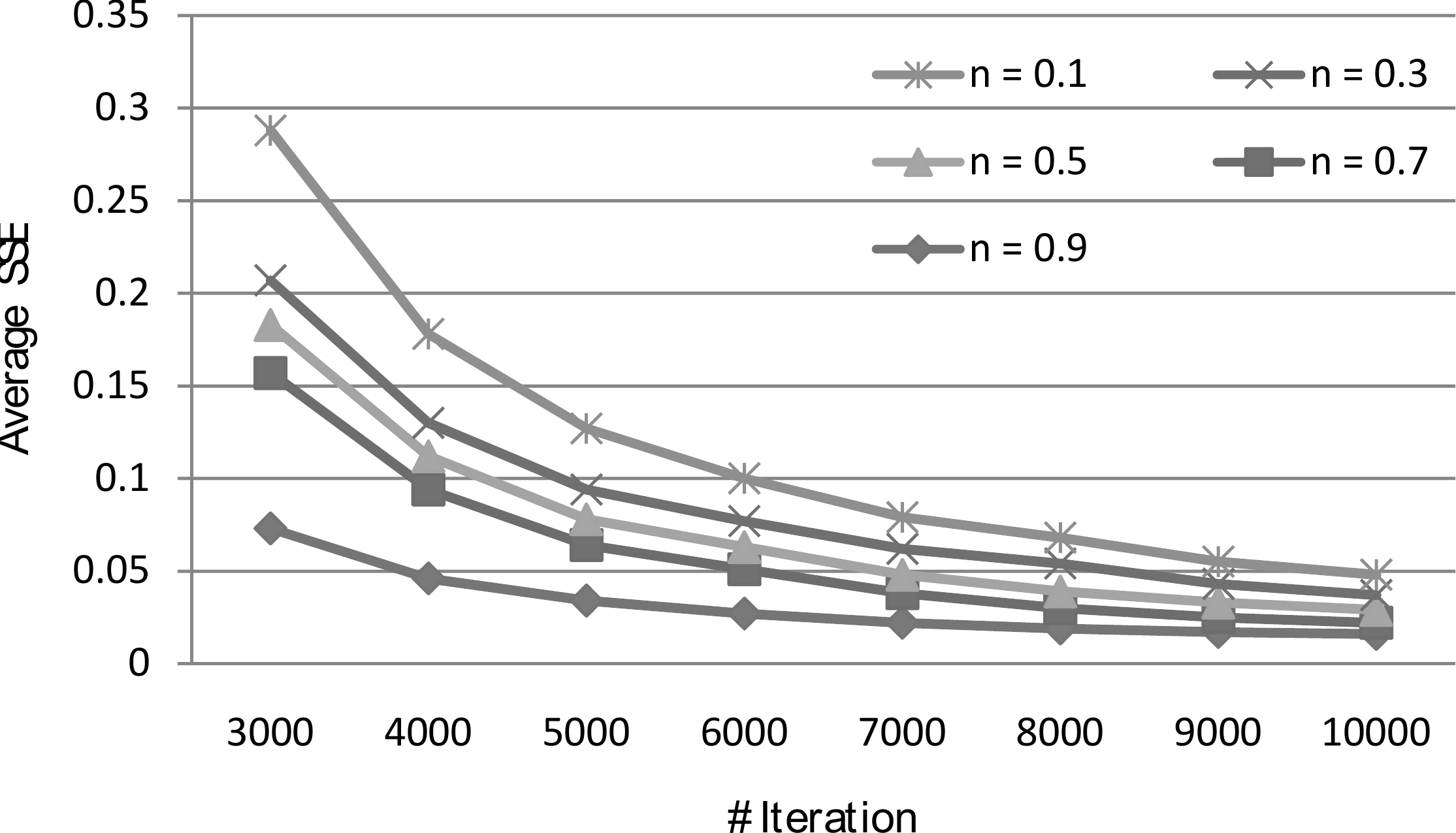} \quad  \includegraphics[width=2in,height=1.5in]{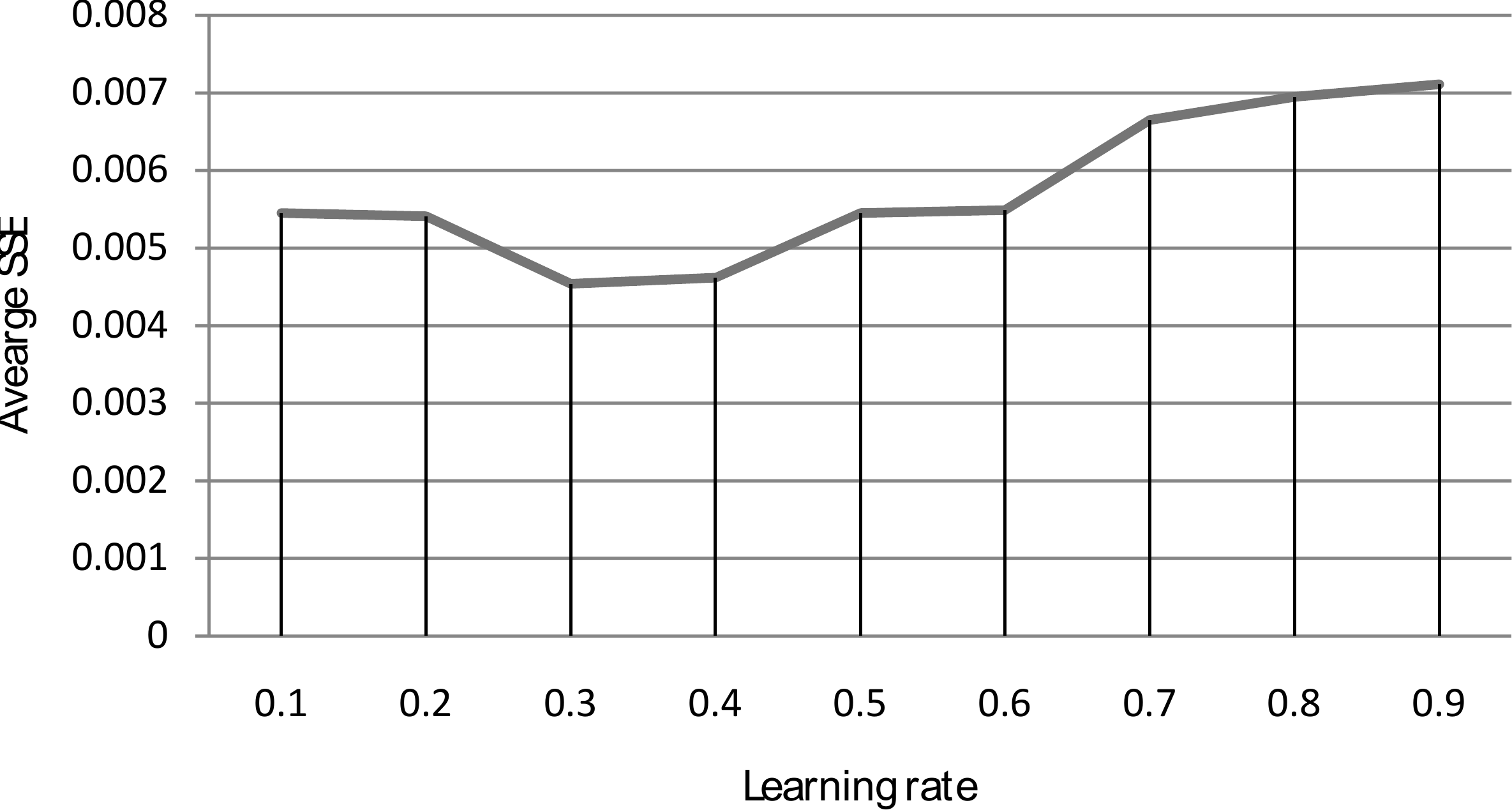}
    \caption{(left) Convergence trajectories at $\eta$; (right) average SSE at various $\eta$}
    \label{fig:learning-a}
    \label{fig:learning-b}
    \vspace{5pt}
    \includegraphics[width=2in,height=1.5in]{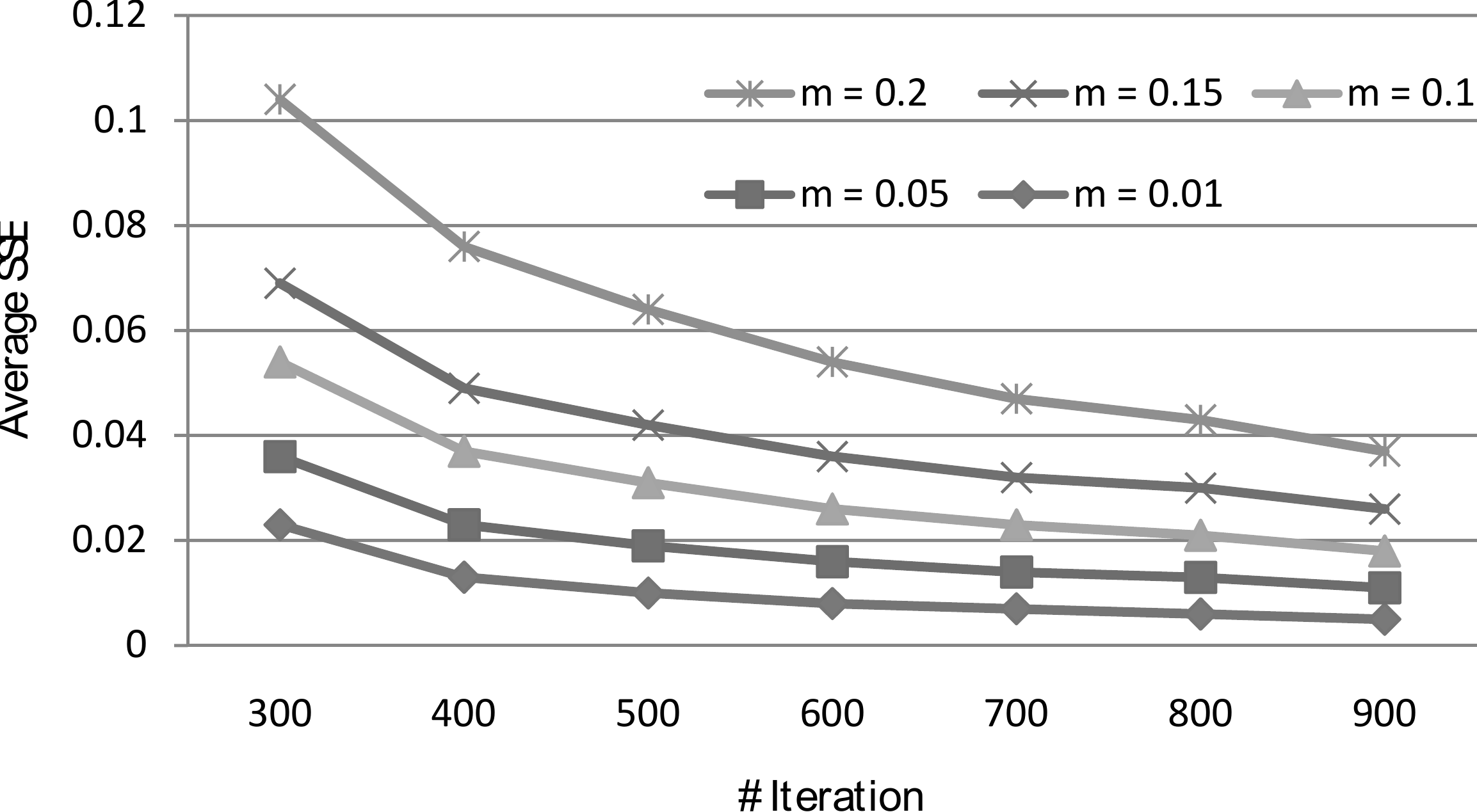} \quad \includegraphics[width=2in,height=1.5in]{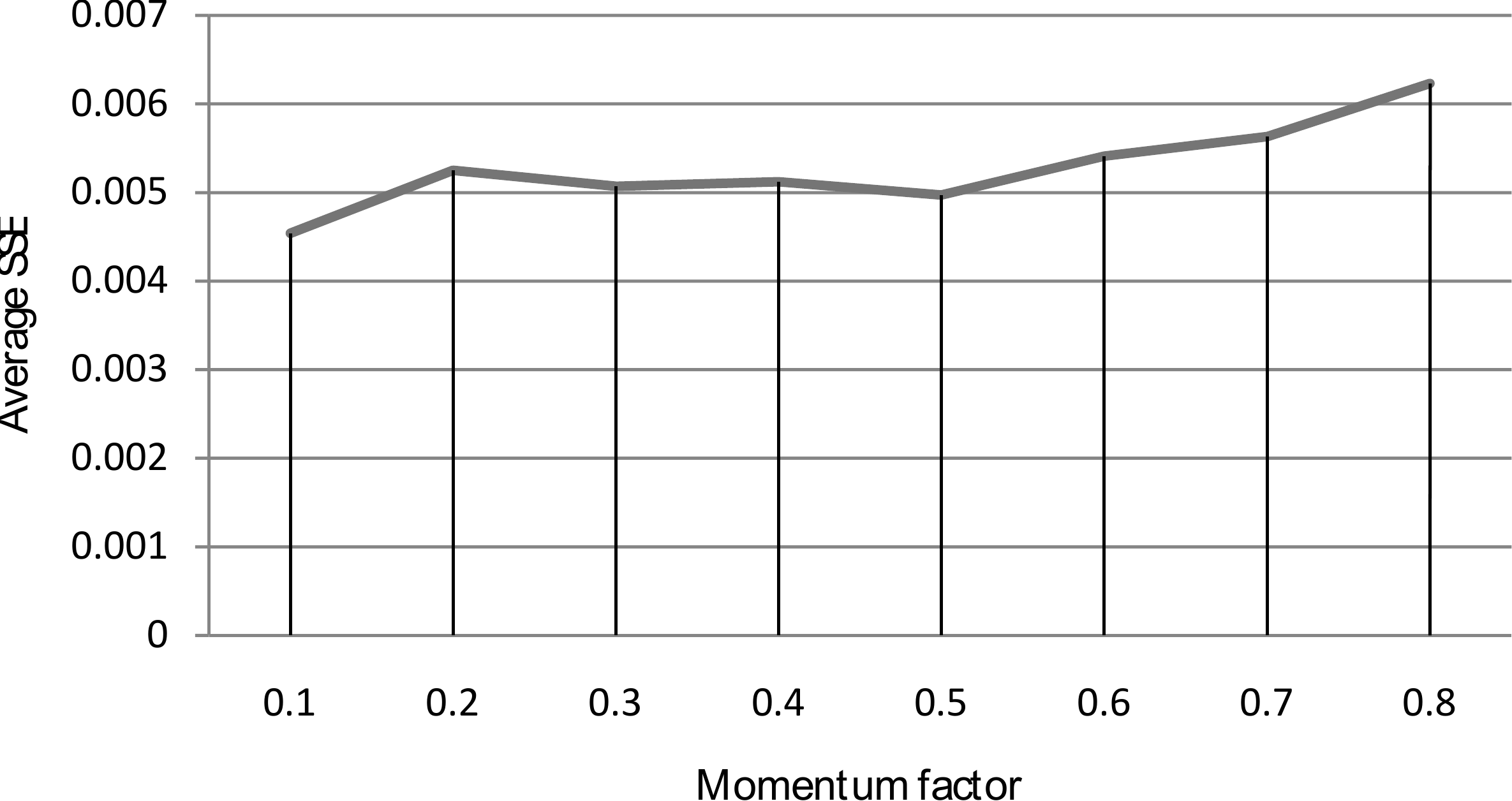}
    \caption{(left) Convergence trajectories at $\beta$; (right) SSE at various $\beta$}
    \label{fig:momentum-a}
    \label{fig:momentum-b}
\end{figure}

Learning rate($\eta$) is crucial for the BP algorithm. It is evident from the BP algorithm mentioned in the subsection \ref{subsec:bpalgo} that the weight changes in BP algorithm are proportional to the derivative of the error. The $\eta$ controls the weight change. With larger $\eta$, the larger being the weight changes in each epoch. Training/Learning of the NN is faster if the $\eta$ is larger and slower if the $\eta$ is lower. The size of the $\eta$ can influence the fact that whether the network achieves a stable solution. A true gradient descent technique should take very little steps to build solution. If the $\eta$ gets too large then the algorithm may disobey gradient descent procedure. Two distinct experiments have been done using $\eta$ and their results are illustrated in Figure \ref{fig:learning-a}. 
In the first experiment as shown in left of Figure \ref{fig:learning-a}, network configuration was set to 5 - 5 - 5, number of iteration was set to 10000, $\beta$ considered was 0.1, training set volume was 200 and batch mode training was adopted. From the experiment, it was observed that for larger values of $\eta$, $\bigtriangleup W_{ij}$ (weight change) was large and for small $\eta$ the $\bigtriangleup W_{ij}$ (weight change) was small. At a fixed iteration 10000 and at $\eta$ 0.9, the network training was fast. We got SSE 0.02 for $\eta$ 0.9 and for $\eta$ 0.1 the SSE was 0.05 that indicated that for the small $\eta$, algorithm required more steps to converge, though the convergence was guaranteed because the small steps due to small $\eta$ minimized the chance of escaping global minima. With larger values of $\eta$, the algorithm may escape the global minima that is clearly indicated in Figure \ref{fig:learning-b}. In second experiment shown in right of Figure \ref{fig:learning-b}, it may be observed that for the lower $\eta$ i.e., 0.1 and 0.2, the algorithm fell short to get reach to a good result in limited number of iteration. However, at $\eta$ 0.3 and 0.4, the algorithm provided  a good result. Whereas, learning rates higher than 0.4 were induced to escape of global optima.

Slight modification to BP weight update rule, additional momentum ($\beta$) term was introduced by Rumelhart in \cite{Rummelhart}. The concept of momentum is that the previous chance in weight should influence the current direction of movement in search space. The term momentum indicates that once the weight starts moving in a particular direction it continues  to move in that direction. Momentum can help in speeding up the learning and can help in escaping local minima. But too much speed may become unhealthy for the training process. 
Figure \ref{fig:momentum-a} indicates that for the lower values of the momentum rate, the algorithm performs well. Therefore, the $\beta$ value should be increased as per learning speed requirement.

\subsubsection{Training Volume}
\label{subsec:training_vol}
We have already mentioned that the performance of BP algorithm depends on its parameters and the complexity of problem. We can regulate those parameters to improve the performance of algorithm but we can not control the complexity of the problem. The performance of the algorithm is highly influenced by the size of training set and the characteristic data within the training set.
\begin{figure}
  \centering
    \includegraphics[width=2in,height=1.5in]{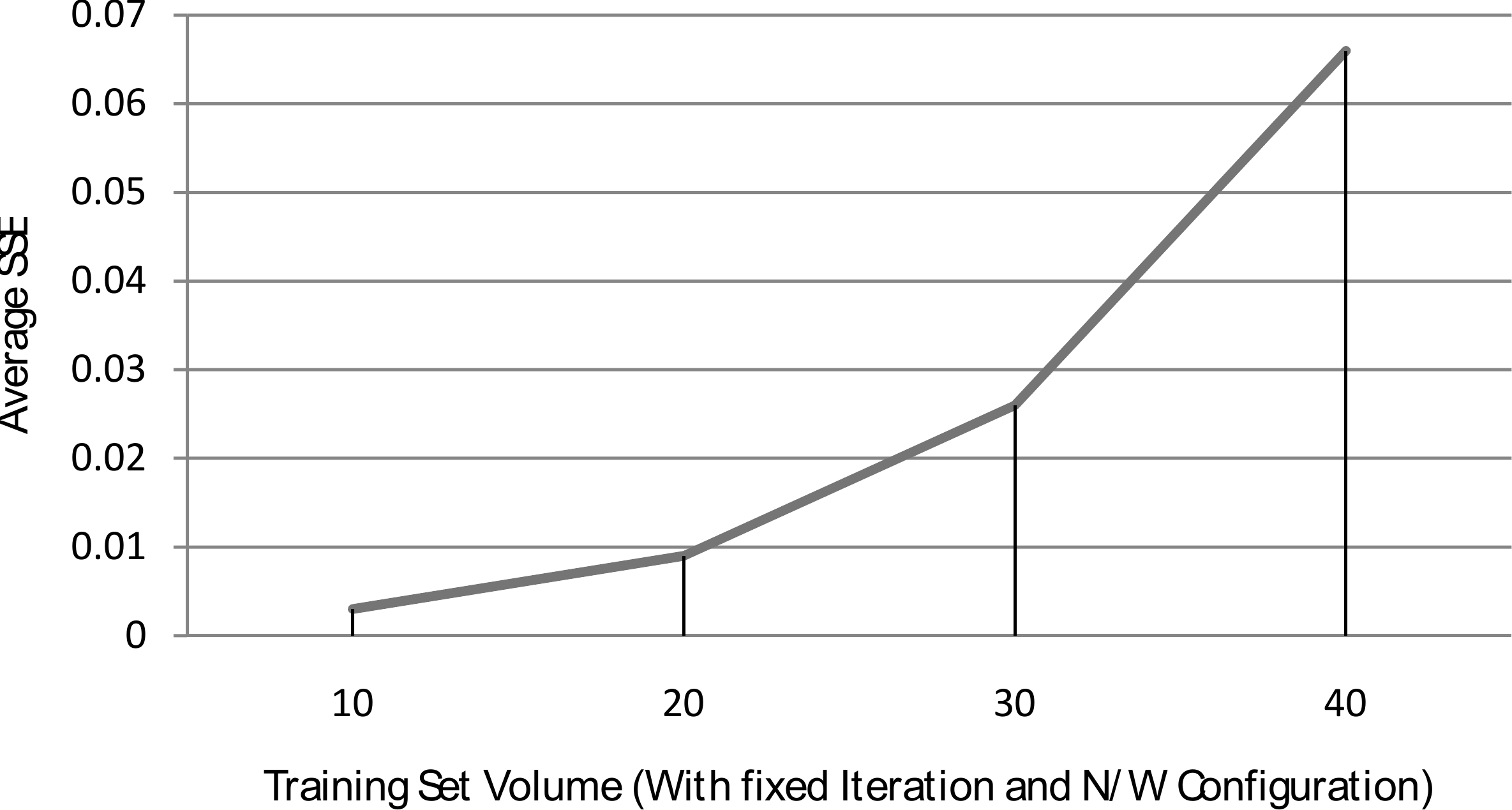} \quad \includegraphics[width=2in,height=1.5in]{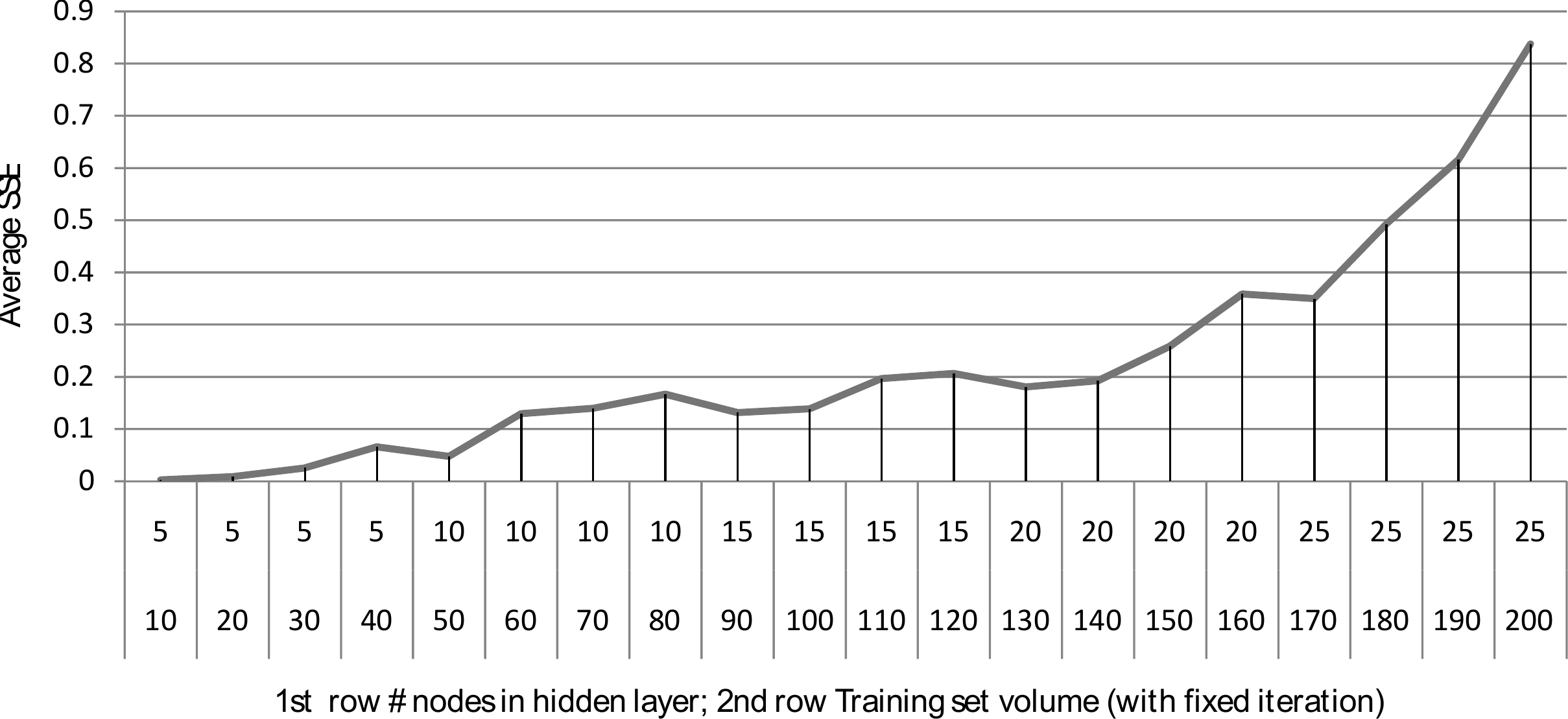}
    \caption{(left) Fixed NN configuration \& iteration (right) Fixed iteration}
    \label{fig:volume-a}
    \label{fig:volume-b}
    \vspace{5pt}
    \includegraphics[width=2in,height=1.5in]{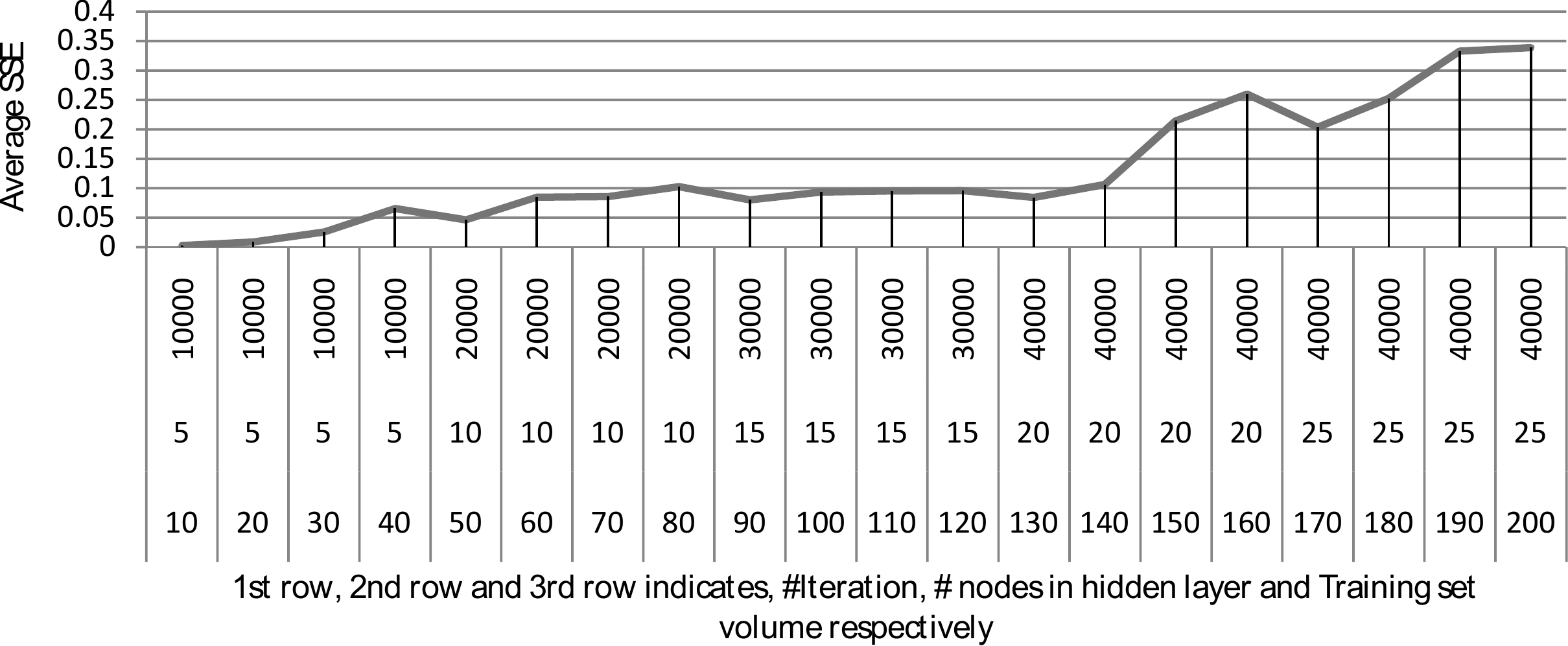} \quad \includegraphics[width=2in,height=1.5in]{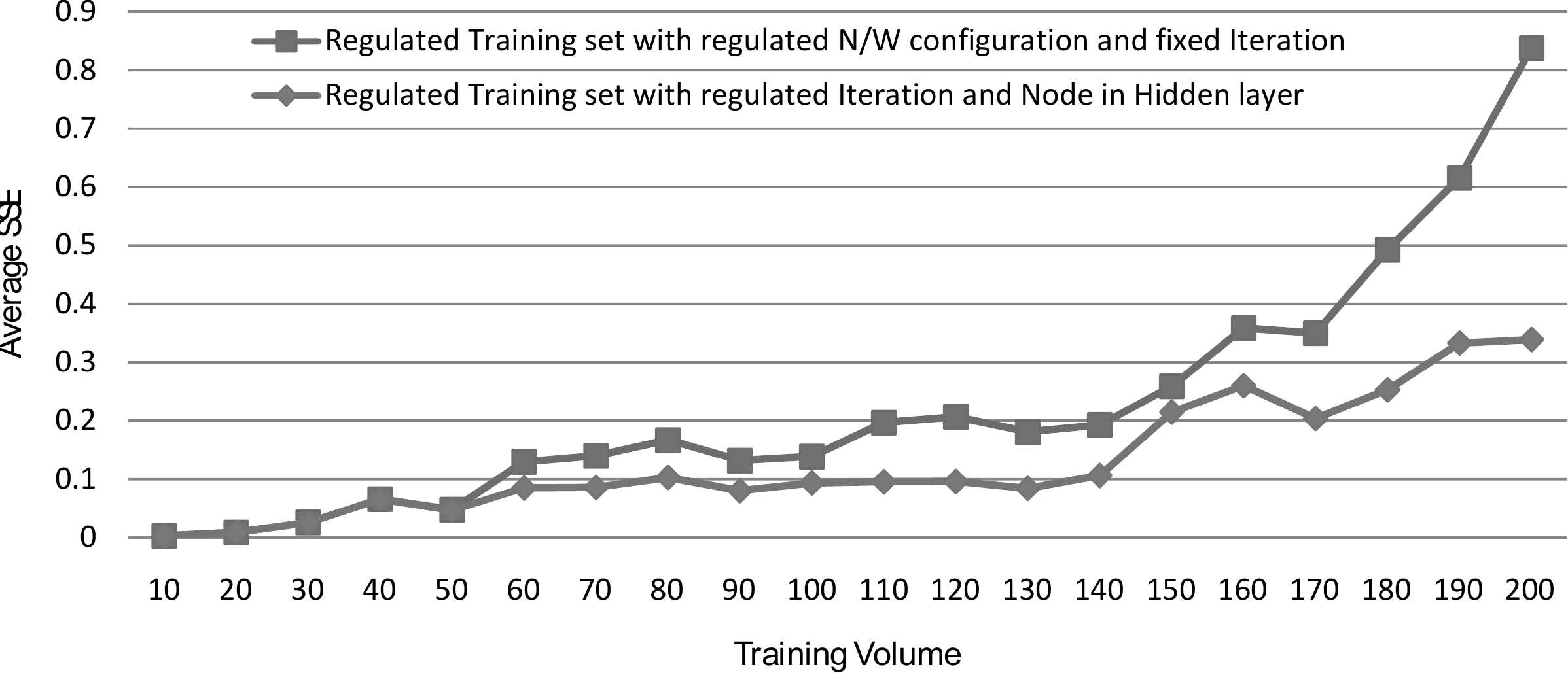}
    \caption{(left) SSE vs. Training volume; (right) Superimposition of figure \ref{fig:volume-c} over \ref{fig:volume-b}}
    \label{fig:volume-c}
    \label{fig:volume-d}
\end{figure}
Figures \ref{fig:volume-a} and \ref{fig:volume-d}, we present an observation based on the volume of training set. In Figure \ref{fig:volume-a} (left), the X-axis indicates the volume of training set whereas the Y-axis indicates the average SSE achieved. Figure \ref{fig:volume-a}(left) indicates that the average SSE has been plotted against different training set volume feeding to networks of fixed size trained at same iteration. From Figure \ref{fig:volume-a} (left), it may be observed that the performance of algorithm gets poorer for the increasing volume of the training set. To improve the performance of the algorithm, we should reconfigure NN to higher configuration and/or increase maximum training iteration values each time the training set volume was increased. In Figure \ref{fig:volume-b} (right), the first row values in the X-axis indicate the number of hidden layer nodes in three layered NN, the second row values in the X-axis indicates the volume of training set whereas, the Y-axis indicates the average SSE achieved. In Figure \ref{fig:volume-c} (left), the first row, second row and third row values in the X-axis indicates the maximum iteration, number of hidden layer nodes in three layered NN and volume of training set respectively while the Y-axis indicates the average SSE achieved. From Figure \ref{fig:volume-b} it may be observed that for the increasing values of training set volume, the average SSE was increases. Hence, the performance became poorer, but as soon as the network was reconfigured to a higher configuration the average SSE dips down. At that particular configuration, when the training set volume was increased, the SSE was also increased. Figure \ref{fig:volume-d} (right) is a superimposition of Figure \ref{fig:volume-c} (left) over Figure \ref{fig:volume-b} (right). Figure \ref{fig:volume-d} (right) indicates that iteration upgradation and network reconfiguration together became necessary for the improvement of algorithm when the training set volume was increased.

\subsection{Complexity Analysis }  
\label{subsec:ca}    
The time complexity of BP neural network algorithm depends on the number of iteration it takes to converge, the number of patterns in the data sample and the time complexity needed to update synaptic weights. Hence, it is clear that the complexity of BP algorithm is problem dependent. Let $n$ iterations are required for the convergence of BP algorithm. Let $p$ be the total number of patterns in the training data. The synaptic weights of the NN shown in Figure \ref{fig:nn} may be represented as a weight vector. Hence to update synaptic weights, the running time complexity required is $O(m)$, where m is size of weight vector. To update synaptic weights, we need to compute gradient as per equations \ref{eq:grad}, computation of gradient for each pattern takes $O(n^2)$. The weights may be updated either in sequential mode or in batch mode. Subsection \ref{subsec:pa} contains detailed discussion on the mode of NN training using BP algorithm. In batch mode, weights are updated once in an epoch. One epoch training means training of NN for entire patterns in the training set. In sequential mode, weights are updated for each pattern presented to the network. Let $w$ be the cost of the gradient computation that is basically $O(m)$, equivalent to cost of updating weights. Whichever the training mode it may be, gradients are computed for each training pattern. In sequential mode, weight updation and gradient computation are parallel process. In batch mode, weight update once in an epoch, whose contribution is feeble in total cost and may be ignored. Hence, the complexity of BP algorithm is stands to $O(p \times w \times n) = O(pwn)$.

\subsection{Comparison with Other NN Training Methods} 
\label{subsec:compare}     
We have adopted four different intelligent techniques for the training of NN. These adopted techniques are namely BP algorithm, conjugate gradient (CG) method, genetic algorithm (GA) \cite{Ojha-ga-j,Ojha-ga-ic} and particle swarm optimization (PSO) algorithm \cite{Ojha-pso-j,Ojha-pso-ic} for the training of NN. The present section offers a comprehensive performance study and comparison between these intelligent techniques applied for manhole gas detection problem. 
\subsubsection{Empirical Analysis}
\begin{figure}[b]
  \centering
    \includegraphics[width=2in,height=1.5in]{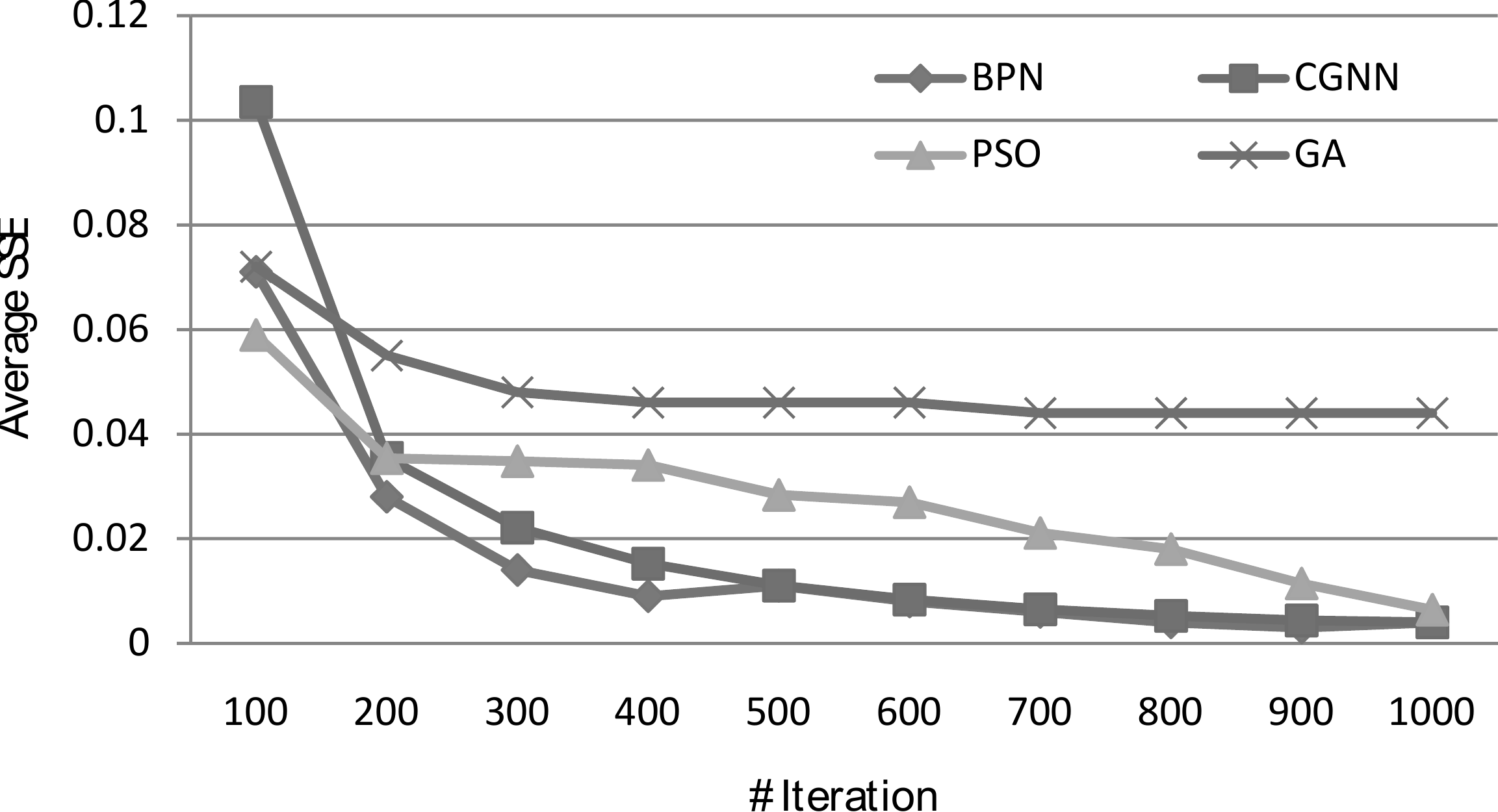} \quad \includegraphics[width=2in,height=1.5in]{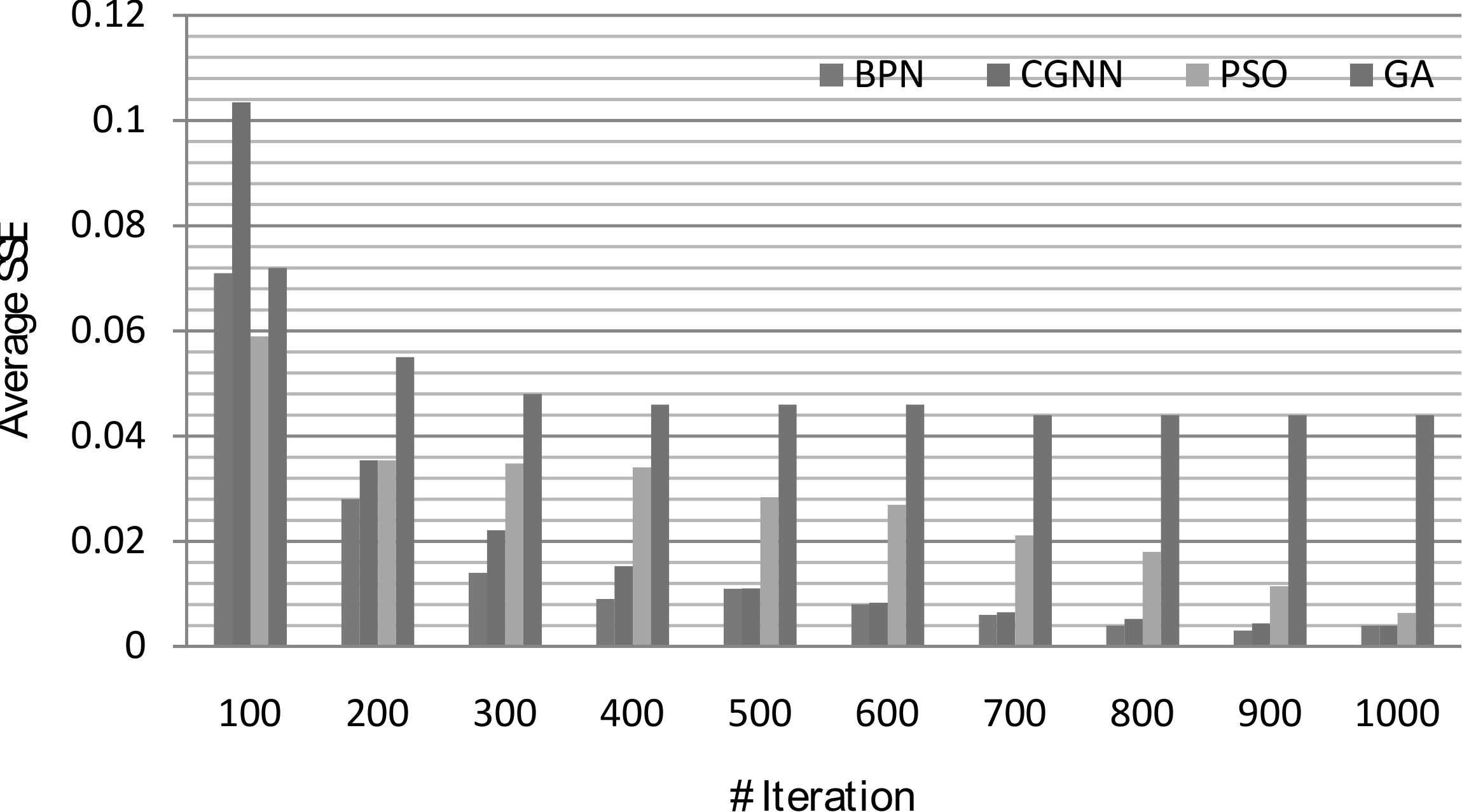}
    \caption{Convergence trajectory analysis; (right) SSE @ various epochs}
    \label{fig:comp-a}
    \label{fig:comp-b}
%
%
%
\end{figure}
In Figure \ref{fig:comp-a}, the X-axis indicates number of iterations while the Y-axis indicates the average SSE achieved against different iterations. Figure \ref{fig:comp-a} (left) indicate the convergence trajectory and \ref{fig:comp-b} (right) indicates SSE obtained in various epochs in training process by mentioned the algorithm. It was observed that the BP algorithm converged faster than the other algorithms. The convergence trajectory of CG method appeared smoother than that of the BP algorithm and its SSE value got reduced to value nearly equal to the value of SSE achieved by the BP algorithm. Although the convergence trajectory of the PSO approach was not as convincing that of the BP algorithm and CG method, it was observed that the PSO approach was efficient enough to ensure the SSE nearer to the one achieved using classical NN training algorithms. Figure \ref{fig:comp-a} indicate that the GA was not as efficient as the other three approaches. GA quickly gets stuck into local optima as far as the present application was concerned. 
\subsubsection{Theoretical Analysis}
The complexity analysis of BP algorithm is provided in subsection \ref{subsec:ca}. The cost met by line search in CG method is an additional computational cost in contrast with BP algorithm counterpart. The computational cost met by PSO and GA algorithm are given as the $O(pqwng)$, where $q$ is size of population, $n$ number of iterations, $p$ is the number of training examples, $w$ is the cost needed to update synaptic weights and $g$ is the cost of producing next generation. It may please be noted that the cost of $g$ in the PSO and the GA are depends on their own dynamics of producing next generation. In GA it is based on selection, crossover, and mutation operation \cite{Goldberg}, where the PSO has simple nonderivative methods of producing next generation \cite{Kennedy}.
 
\subsubsection{Statistical Analysis}
The Kolmogorov-Smirnov test (KS-test) being nonparametric in nature does not make any assumption about the distribution of data. The two-sample KS test is useful for comparing two samples, as it is sensitive to differences in both location and shape of the empirical cumulative distribution functions ($epcdf$) of two samples. Clearly speaking, KS-test tries to determine if two datasets $X$ and $Y$ differ significantly \cite{pdthesis}. The KS-test make the following hypothesis.
The null hypothesis $H_0$ indicates that the two underlying one dimensional unknown probability distributions corresponding to $X$ and $Y$ are indistinguishable i.e. datasets $X$ and $Y$ are statistically similar ($X \equiv Y$). The alternative hypothesis $H^t_1$ indicates that $X$ and $Y$ are distinguishable i.e datasets $X$ and $Y$ are statistically dissimilar. If it is the alternative hypothesis than their order (direction) becomes an important consideration. We need to determine whether the former (dataset $X$) is stochastically larger than or smaller than the later one (dataset $Y$) i.e. $X \succ Y$ or $X \prec Y$. Such KS test is known as one sided test and the direction is determined by the distance $D^+_{n,m}$, $D^-_{n,m}$ and $D^{ }_{n,m}$, where, value $n$ and $m$ are the cardinality of the set $X$ and $Y$ respectively. The null hypothesis $H_0$ is rejected if $D^{ }_{n,m} > K_{\alpha}$, where $K_{\alpha}$ is critical value \cite{Critical_Value}. For $n$ and $m$ being 20 samples size, $K_{\alpha}$ was found to be equal to 0.4301 for $alpha$ = 0.05. The vale of $\alpha$ indicates 95\% confidence in the test. Readers may explore \cite{Critical_Value,KSTest,KSTest1} to mitigate their more interest in KS test. To perform KS test, we took  20 instances of SSE produced by each algorithm applied on the given problem. KS test was conducted in between BP and  other intelligent algorithms. Hence, set $X$ was prepared with the SSEs of BP and three separate sets of $Y$ was prepared using the SSE values of CG, PSO and GA algorithm. The outcome of the KS test is provided in Table\ref{tab:kstest} that itself is conclusive about the significance of the BP algorithm.      
   
\begin{table}[t]
\centering
\caption{KS test: BP vs. Other Intelligent Algorithms}
\label{tab:kstest}
    \setlength{\tabcolsep}{2pt}
	\small
	\begin{tabular}{ c | c | c | c }
	\hline
	\textbf{KS Test type} & \multicolumn{3}{c}{\centering \textbf{Intelligent Algorithms ($Y$)}}\\
	\cline{2-4}
BP ($X$)
	& \multicolumn{1}{c|}{CGNN}
	& \multicolumn{1}{c|}{PSO}
	& \multicolumn{1}{c}{GA}\\
	\hline
    D$^{+}_{nm}$ & 0.10 & 0.15 & 1.00 \\
	D$^{-}_{nm}$ & 0.20 & 0.15 & 0.00\\
	D$^{ }_{nm}$ & 0.20 & 0.15 & 1.00\\
	\hline 
	Decision & $X \equiv Y$ & $X \equiv Y$ & $X \succ Y$\\
\hline 
\end{tabular}
\end{table}

\section{Results and Discussion}
\label{sec:res}
The data sample was prepared as per the procedures mentioned in section \ref{subsec:datacollection}. Collected data sample was partitioned in two sets. About eighty percent of the original set was used as training set, remaining twenty percent was used for testing purpose. After training, the system undergone for the test using the test set. The output of the trained NN was denormalized in order to present output in terms of concentration of the gas components present in the given test sample/gaseous mixture. We are providing a sample test result obtained for the input sample 2, which is provided in Table \ref{tab:trainingset}. The predicted concentration value corresponding to the given input sample is shown in Table \ref{tab:result}. In table \ref{tab:result}, the interpretation column is based on the comparison between the denormalized value of NN output and safety limits of the respective gases. Note that each of the nodes at the output layer dedicated to a particular gas. Safety limit of the manhole gases are as follows. Safety limit of $NH_{3}$ is laying between 25 - 40ppm (as per limit set by World Health Organization), $CO$ is in between 35 - 100ppm \cite{CO,CO1,CO2}, $H_{2}S$ is in between 50 - 100ppm \cite{H2S,H2S1}, $CO_{2}$ is in between 5000 - 8000ppm \cite{CO2,CO21} and $CH_{4}$ is on between 5000 - 10000 ppm \cite{CH4}. 
\begin{table}
\begin{center}
\caption{System result presentation in ppm}
\label{tab:result}
\begin{tabular}{c | c | c | c | l | c}
\hline
Input Gas & \multicolumn{3}{| c |}{ Responding unit} & \multicolumn{1}{| c |}{ Safety limit }& Interpretation \\
\cline{2-4}
          & Sensor & NN & System &  \\ 
\hline
$NH_{3}$ &	0.260 &	0.016 &	\,\, \, 80 ppm & 25 - 40ppm       & Unsafe\\
$CO $    &	0.346 &	0.022 &	\, 110 ppm     & 35 - 100ppm      & Unsafe\\
$H_{2}S$ &	0.240 &	0.023 &	\, 115 ppm     & 50 - 100ppm      & Unsafe\\
$CO_{2}$ &	0.142 &	0.022 &	\, 110 ppm     & 5000 - 8000ppm   & Safe\\
$CH_{4}$ &	0.843 &	0.993 &	4965 ppm       & 5000 - 10000 ppm & Safe\\
\hline 
\end{tabular} 
\end{center}
\end{table}

\section{Conclusion}
\label{Conclusions}
In present chapter, we have discuss the design issues of an intelligent sensory system (ISS) comprising semiconductor based GSA and NN regressor. The BP was employed for the supervised training of the NN model developed. The proposed design of ISS offered solution to manhole gas mixture detection. The problem was vied as noise reduction/pattern recognition problem. We have discussed the mechanisms involved in preparation and collection of data sample for ISS. The significant issues of cross sensitivity was firmly addressed in this chapter. We have discussed the issues in training of NN using backprpoagation (BP) algorithm. A comprehensive performance study of BP as supervised NN training algorithm was provided in this chapter. Performance of BP was meticulously analyzed for real life application problem. Performance comparison in terms of empirical, theoretical and statistical sense between the BP and various other hybrid intelligent approaches applied on the said problem was provided in this chapter. A concise discussion on the safety limits and system result presentation mechanism was presented in remainder section of the chapter. The data sample in the present problem may not represent the entire spectrum of the problem. Therefore, at present it was a non-trivial task for the NN regresor. Hence, it offered a high quality results. Therefore, an interesting study over larger dataset to examine how the manhole gas detection problem can be framed as a classification problem using the available classifier tools.   

\begin{acknowledgement}
This work was supported by Department of Science \& Technology (Govt. of India) for the financial supports vide Project No.: IDP/IND/02/2009 and the IPROCOM Marie Curie initial training network, funded through the People Programme (Marie Curie Actions) of the European Union's Seventh Framework Programme FP7/2007-2013/ under REA grant agreement No. 316555.
\end{acknowledgement}

\bibliographystyle{IEEEtran}  
\bibliography{masterBibTex}   

\begin{thebibliography}{10}
\providecommand{\url}[1]{#1}
\csname url@samestyle\endcsname
\providecommand{\newblock}{\relax}
\providecommand{\bibinfo}[2]{#2}
\providecommand{\BIBentrySTDinterwordspacing}{\spaceskip=0pt\relax}
\providecommand{\BIBentryALTinterwordstretchfactor}{4}
\providecommand{\BIBentryALTinterwordspacing}{\spaceskip=\fontdimen2\font plus
\BIBentryALTinterwordstretchfactor\fontdimen3\font minus
  \fontdimen4\font\relax}
\providecommand{\BIBforeignlanguage}[2]{{%
\expandafter\ifx\csname l@#1\endcsname\relax
\typeout{** WARNING: IEEEtran.bst: No hyphenation pattern has been}%
\typeout{** loaded for the language `#1'. Using the pattern for}%
\typeout{** the default language instead.}%
\else
\language=\csname l@#1\endcsname
\fi
#2}}
\providecommand{\BIBdecl}{\relax}
\BIBdecl

\bibitem{SewerGas}
J.~Whorton, ``The insidious foe"--sewer gas",'' \emph{West. J. Med.}, vol. 175,
  no.~6, p. 427–428, Decemberl 2001, pMC 1275984.

\bibitem{Lewis}
Lewis, \emph{Dangerous Properties of Industrial Materials}, ninth
  edition~ed.\hskip 1em plus 0.5em minus 0.4em\relax Van Nostrand Reinhold,
  1996, iSBN 0132047674.

\bibitem{SewerGas1}
N.~Gromicko, ``Sewer gases in the home,'' 2006,
  http://www.nachi.org/sewer-gases-home.html.

\bibitem{NIOSH}
NIOSH, ``Volunteer fire fighter dies during attempted rescue of utility worker
  from a confined space,'' 2011,
  http://www.cdc.gov/niosh/fire/reports/face201031.html.

\bibitem{theHinduMarch2014}
T.~Hindu, ``Supreme court orders states to abolish manual scavenging,'' March
  2014,
  http://www.thehindu.com/news/national/supreme-court-orders-states-to-abolish-manual-scavenging/article5840086.ece.

\bibitem{toiMarch2014}
T.~T. of~India, ``Sewer deaths,'' March 2014,
  http://timesofindia.indiatimes.com/city/delhi/Panel-holds-DDA-guilty-for-sewer-death/articleshow/31916051.cms.

\bibitem{theHinduApril2014}
T.~Hindu, ``Deaths in the drains,'' April 2014,
  http://www.thehindu.com/opinion/op-ed/deaths-in-the-drains/article5868090.ece?homepage=true.

\bibitem{theHinduApril52014}
------, ``Sewer deaths,'' April 2014,
  http://www.thehindu.com/opinion/letters/sewer-deaths/article5873493.ece.

\bibitem{Jun}
J.~Li, ``A mixed gas sensor system based on thin film saw sensor array and
  neural network.''\hskip 1em plus 0.5em minus 0.4em\relax IEEE, 1993, pp.
  179--181, 0-7803-0976-6/93.

\bibitem{Sirvastava-ga}
A.~K. Srivastava, S.~K. Srivastava, and K.~K. Shukla, ``In search of a good
  neuro-genetic computational paradigm.''\hskip 1em plus 0.5em minus
  0.4em\relax IEEE, 2000, pp. 497--502, 0-78O3-5812-0/00.

\bibitem{Sirvastava-enose}
------, ``On the design issue of intelligent electronic nose system.''\hskip
  1em plus 0.5em minus 0.4em\relax IEEE, 2000, pp. 243--248, 0-7803-5812-0/00.

\bibitem{Eduard}
E.~L. et.al., ``Multicomponent gas mixture analysis using a single tin oxide
  sensor and dynamic pattern recognition,'' \emph{IEEE SENSORS JOURNAL}, vol.
  Vol. 1, no. No.3, pp. 207--213, October 2001, 1530�437X/01.

\bibitem{Junhua-Liu}
J.~Liu, Y.~Zhang, Y.~Zhang, and M.~Cheng, ``Cross sensitivity reduction of gas
  sensors using genetic algorithm neural network,'' in \emph{Optical Methods
  for Industrial Processes}, S.~Farquharson, Ed., vol. Vol. 4201.\hskip 1em
  plus 0.5em minus 0.4em\relax Proceedings of SPIE, 2001.

\bibitem{Georgios}
G.~Tsirigotis, L.~Berry, and M.~Gatzioni, ``Neural network based recognition,
  of $co$ and $nh_3$ reducing gases, using a metallic oxide gas sensor array,''
  in \emph{Scientific Proceedings of RTU}, ser. Series 7, vol. Vol. 3.\hskip
  1em plus 0.5em minus 0.4em\relax Telecommunications and Electronics, 2003,
  pp. 6--10.

\bibitem{Dae}
D.-S. Lee, S.-W. Ban, M.~Lee, and D.-D. Lee, ``Micro gas sensor array with
  neural network for recognizing combustible leakage gases,'' \emph{IEEE
  SENSORS JOURNAL}, vol. Vol 5, no. No. 3, pp. 530--536, June 2005, dOI
  10.1109/JSEN.2005.845186.

\bibitem{Maxim}
M.~Ambard, B.~Guo, D.~Martinez, and A.~Bermak, ``A spiking neural network for
  gas discrimination using a tin oxide sensor array,'' in \emph{International
  Symposium on Electronic Design, Test \& Applications}.\hskip 1em plus 0.5em
  minus 0.4em\relax IEEE, 2008, pp. 394--397, 0-7695-3110-5/08.

\bibitem{Hakim}
H.~Baha and Z.~Dibi, ``A novel neural network-based technique for smart gas
  sensors operating in a dynamic environment,'' \emph{sensors}, vol. Vol. 9,
  pp. 8944--8960, 2009, iSSN 1424-8220.

\bibitem{Wu}
W.~Pan, N.~Li, and P.~Liu, ``Application of electronic nose in gas mixture
  quantitative detection,'' in \emph{Proceedings of IC-NIDC}.\hskip 1em plus
  0.5em minus 0.4em\relax IEEE, 2009, pp. 976--980, 978-1-4244-4900-2/09.

\bibitem{Chatchawal}
C.~Wongchoosuka, A.~Wisitsoraatb, A.~Tuantranontb, and T.~Kerdcharoena,
  ``Portable electronic nose based on carbon nanotube-$sno_2$ gas sensors and
  its application for detection of methanol contamination in whiskeys,''
  \emph{Sensors and Actuators B: Chemical}, 2010,
  doi:10.1016/j.snb.2010.03.072.

\bibitem{Qian}
Q.~Zhang, H.~Li, and Z.~Tang, ``Knowledge-based genetic algorithms data fusion
  and its application in mine mixed-gas detection.''\hskip 1em plus 0.5em minus
  0.4em\relax IEEE, 2010, pp. 1334--1338, 978-1-4244-5182-1/10.

\bibitem{Won}
D.~S. E. S.~Y. Won~So, Jamin~Koo, ``The estimation of hazardous gas release
  rate using optical sensor and neural network,'' in \emph{European Symposium
  on Computer Aided Process Engineering � ESCAPE20}, S.~Pierucci and G.~B.
  Ferraris, Eds.\hskip 1em plus 0.5em minus 0.4em\relax Elsevier B.V, 2010.

\bibitem{Ojha-bp-j1}
V.~K. Ojha, P.~Duta, H.~Saha, and S.~Ghosh, ``Detection of proportion of
  different gas components present in manhole gas mixture using backpropagation
  neural network,'' \emph{International Proceedings Of Computer Science \&
  Information Technology}, vol. Vol 1, pp. 11--15, April 2012, iSBN:
  978-981-07-2068-1.

\bibitem{Ojha-lr-ic}
------, ``Linear regression based statistical approach for detecting proportion
  of component gases in manhole gas mixture,'' in \emph{In International
  Symposium On Physics And Technology Of Sensors}.\hskip 1em plus 0.5em minus
  0.4em\relax IEEE, March 2012, dOI:10.1109/ISPTS.2012.6260865.

\bibitem{Ojha_sa_ic}
------, ``A novel neuro simulated annealing algorithm for detecting proportion
  of component gases in manhole gas mixture,'' in \emph{International
  Conference on Advances in Computing and Communications}.\hskip 1em plus 0.5em
  minus 0.4em\relax IEEE, August 2012, pp. 238--241, dOI 10.1109/ICACC.2012.54.

\bibitem{GSA}
S.~Ghosh, A.~Roy, S.~Singh, V.~K. Ojha, P.~Dutta, and H.~Saha, ``Sensor array
  for manhole gas analysis,'' in \emph{International Symposium on Physics and
  Technology of Sensors.}\hskip 1em plus 0.5em minus 0.4em\relax Pune,India:
  IEEE, March 2012, 978-1-4673-1040-6.

\bibitem{PortableGSA}
S.~Ghosh, C.~Roychaudhuri, H.~Saha, V.~K. Ojha, and P.~Dutta, ``Portable sensor
  array system for intelligent recognizer of manhole gas,'' in
  \emph{International Conference On Sensing Technology (ICST)}.\hskip 1em plus
  0.5em minus 0.4em\relax Pune,India: IEEE, December 2012, pp. 589 -- 594,
  978-1-4673-2246-1.

\bibitem{Simon}
S.~Haykin, \emph{Neural Networks A Comprehensive Foundation}, second
  edition~ed.\hskip 1em plus 0.5em minus 0.4em\relax Pearson Prentice Hall,
  2005, iSBN 81-7803-300-0.

\bibitem{Rummelhart}
D.~E. Rummelhart, G.~E. Hinton, and R.~J. Williams, ``Learning representations
  by back-propagating errors,'' \emph{Nature}, vol. 323, no. 6088, p.
  533�536, October 1986, doi:10.1038/323533a0.

\bibitem{Sivanadam}
S.~N. Sivanadam and S.~N. Deepa, \emph{Principles of Soft Computing}.\hskip 1em
  plus 0.5em minus 0.4em\relax Wiley India (p) Ltd, 2007, iSBN: 81-265-1075-7.

\bibitem{Ojha-ga-j}
V.~K. Ojha, P.~Duta, and H.~Saha, ``Performance analysis of neuro genetic
  algorithm applied on detecting proportion of components in manhole gas
  mixture,'' \emph{International Journal Of Artificial Intelligence \&
  Application}, vol. Vol 3, no. No. 4, pp. 83--98, July 2012, dOI :
  10.5121/IJAIA.2012.3406.

\bibitem{Ojha-ga-ic}
V.~K. Ojha, P.~Duta, H.~Saha, and S.~Ghosh, ``Application of real valued neuro
  genetic algorithm in detection of components present in manhole gas
  mixture,'' in \emph{Advances In Intelligent And Soft Computing}, D.~C. Wyld,
  Ed., vol. Vol 166.\hskip 1em plus 0.5em minus 0.4em\relax Springer, 2012, pp.
  333--340, dOI:10.1007/978-3-642-30157-5.

\bibitem{Ojha-pso-j}
V.~K. Ojha and P.~Duta, ``Performance analysis of neuro swarm optimization
  algorithm applied on detecting proportion of components in manhole gas
  mixture,'' \emph{Artificial Intelligence Research}, vol. Vol 1, pp. 31--46,
  September 2012, dOI: 10.5430/JNEP.V1N1PX.

\bibitem{Ojha-pso-ic}
V.~K. Ojha, P.~Duta, H.~Saha, and S.~Ghosh, ``A neuro-swarm technique for the
  detection of proportion of components in manhole gas mixture,'' in \emph{In
  Proceedings Of International Conference On Modeling, Optimization And
  Computing.}, vol. Vol 2.\hskip 1em plus 0.5em minus 0.4em\relax
  Kanyakumari,India: NI University, April 2012, pp. 1211--1218.

\bibitem{Goldberg}
D.~E. Goldberg, \emph{Genetic Algorithms in search, Optimization \& Machine
  learning}, first edition~ed.\hskip 1em plus 0.5em minus 0.4em\relax Pearson
  Education, 2006, iSBN 81-7758-829-X.

\bibitem{Kennedy}
J.~Kennedy and R.~C. Eberhart, \emph{Swarm Intelligence}.\hskip 1em plus 0.5em
  minus 0.4em\relax Morgan Kaufmann Publishers, 2001, iSBN: 1-55860-595-9.

\bibitem{pdthesis}
P.~Dutta and D.~DuttaMajumder, \emph{Performance Analysis of Evolutionary
  Algorithm}.\hskip 1em plus 0.5em minus 0.4em\relax Lambert Academic
  Publishers, 2012, iSBN: 978-3-659-18349-2.

\bibitem{Critical_Value}
M.~H. Gail and S.~B. Green, ``Critical values for the one-sided two-sample
  kolmogorov-smirnov statistic,'' \emph{Journal of the American Statistical
  Association}, vol.~71, no. 355, pp. 757--760, September 1976.

\bibitem{KSTest}
D.~C. Boes, F.~A. Graybill, and A.~M. Mood, \emph{Introduction to the Theory of
  Statistics}, 3rd~ed.\hskip 1em plus 0.5em minus 0.4em\relax New York:
  McGraw-Hill, 1974.

\bibitem{KSTest1}
D.~J. Sheskin, \emph{Handbook of Parametric and Nonparametric Statistical
  Procedures}, 3rd~ed.\hskip 1em plus 0.5em minus 0.4em\relax CRC-Press, 2003,
  iSBN-1-58488-440-1.

\bibitem{CO}
A.~Ernst and J.~D. Zibrak, ``Carbon monoxide poisoning,'' \emph{The New England
  Journal of Medicine}, vol. 339, no.~22, November 1998, pMID9828249.

\bibitem{CO1}
M.~Goldstein, ``Carbon monoxide poisoning,'' \emph{Journal of Emergency
  Nursing: JEN: Official Publication of the Emergency Department Nurses
  Association}, 2008, pMID 19022078.

\bibitem{CO2}
D.~Friedman, ``Toxicity of carbon dioxide gas exposure, co2 poisoning symptoms,
  carbon dioxide exposure limits, and links to toxic gas testing procedures,''
  inspectAPedia.

\bibitem{H2S}
USEPA, ``Health and environmental effects problem for hydrogen sulfide,''
  \emph{West. J. Med.}, pp. 118--8, 1980, eCAOCIN026A.

\bibitem{H2S1}
Zenz and O.~D. amd E.P.~Horvath, \emph{Occupational Medicine}, 3rd~ed., 1994,
  p. 886.

\bibitem{CO21}
G.~Shilpa, ``New insight into panic attacks: Carbon dioxide is the culprit,''
  November 2007, inspectAPedia.

\bibitem{CH4}
D.~Fahey, ``Twenty questions and answers about the ozone layer,'' 2002.

\end{thebibliography}

\end{document}